    \PassOptionsToPackage{numbers, compress}{natbib}
\documentclass{article}

\usepackage[ruled,noend,algo2e, linesnumbered]{algorithm2e}

\usepackage[preprint]{neurips_2025}

\usepackage[utf8]{inputenc} %
\usepackage[T1]{fontenc}    %
\usepackage{hyperref}       %
\usepackage{url}            %
\usepackage{booktabs}       %
\usepackage{amsfonts}       %
\usepackage{nicefrac}       %
\usepackage{microtype}      %
\usepackage{xcolor}         %

\usepackage[english]{babel}

\usepackage{amsmath}
\usepackage{graphicx}
\PassOptionsToPackage{dvipsnames}{xcolor}
\usepackage[capitalize, noabbrev, nameinlink]{cleveref}
\usepackage{booktabs}

\usepackage{enumitem}
\usepackage{todonotes}
\usepackage{multirow,multicol}
\definecolor{martin}{rgb}{0.7, 0.2, 0.5}

\usepackage{natbib}
\usepackage{tabularx}
\usepackage{algorithm}
\usepackage{algpseudocode}
\usepackage{placeins} %
\usepackage{subcaption}
\usepackage{colortbl} %
\usepackage{soul}
\SetAlCapHSkip{0pt}
\usepackage{wrapfig}

\SetCommentSty{mycommfont}
\definecolor{darkblue}{cmyk}{1, 0.5, 0, 0}
\definecolor{darkgreen}{cmyk}{.83, .31, .68, .12}
\definecolor{darkred}{cmyk}{.22, 97, .66, .09}
\colorlet{lightblue}{darkblue!50} 
\definecolor{lightgreen}{RGB}{144,238,144}
\crefname{appendix}{App.}{Apps.}
\crefname{figure}{Fig.}{Figs.}
\crefname{section}{Sec.}{Secs.}
\crefname{algorithm}{Alg.}{Alg.}

\DontPrintSemicolon

\title{Hierarchical clustering with maximum density paths and mixture models}

\author{
    Martin Ritzert$^{\star,1}$ 
    \quad 
    Polina Turishcheva$^{\star,1}$
    \quad
    Laura Hansel$^{\star,1}$
    \\
    \textbf{Paul Wollenhaupt}$^{1}$
    \quad
    \textbf{Marissa A. Weis}$^{1}$
    \quad
    \textbf{Alexander S. Ecker}$^{1,2}$
    \medskip
    \\
    $^\star$ Equal Contribution\smallskip\\
    $^1$ Institute of Computer Science and Campus Institute Data Science, University of Göttingen,\\ Göttingen, Germany.
    \\
    $^2$ Max Planck Institute for Dynamics and Self-Organization, Göttingen, Germany.
    \medskip
    \\
    \texttt{\{martin.ritzert,turishcheva,hansel,ecker\}@cs.uni-goettingen.de}\\
}

\begin{document}

\maketitle

\begin{abstract}
Hierarchical clustering is an effective, interpretable method for analyzing structure in data.
It reveals insights at multiple scales without requiring a predefined number of clusters and captures nested patterns and subtle relationships, which are often missed by flat clustering approaches.
However, existing hierarchical clustering methods struggle with high-dimensional data, especially when there are no clear density gaps between modes. 
In this work, we introduce t-NEB, a probabilistically grounded hierarchical clustering method, which yields state-of-the-art clustering performance on naturalistic high-dimensional data.
t-NEB consists of three steps: (1) density estimation via overclustering; (2) finding maximum density paths between clusters; (3) creating a hierarchical structure via bottom-up cluster merging.
t-NEB uses a probabilistic parametric density model for both overclustering and cluster merging, which yields both high clustering performance and a meaningful hierarchy,
making it a valuable tool for exploratory data analysis. 
Code is available at \url{https://anonymous.4open.science/r/tneb_clustering-4056/}. %
\end{abstract} 
\section{Introduction}
Data in the natural sciences is often complex, noisy, and high-dimensional. To find underlying structure, a natural choice is to use clustering techniques.
Hierarchical clustering provides an interpretable, multi-scale view of the data, making it especially useful when the number of clusters is unknown. %
This is often the case in exploratory analyses of real-world datasets, for example when analyzing biological data. 
Modern cluster validity indices (CVIs), which focus on the relation of within- and between-cluster variance, often fail to identify the optimal number of clusters \cite{gagolewski2021cluster}. 
This challenge is amplified in high-dimensional settings, where clusters often partially overlap and lack clear density gaps \citep{tomavsev2016clustering}. 
Clustering often requires capturing multiple levels of granularity such as cell types and their subtypes \citep{harris2018classes, scala2021phenotypic} in neurobiology, emphasizing that the hierarchical structure can be as important as the clustering itself \citep{zeng2022cell}.

\begin{figure}[t]
    \centering
    \includegraphics[width=\linewidth]{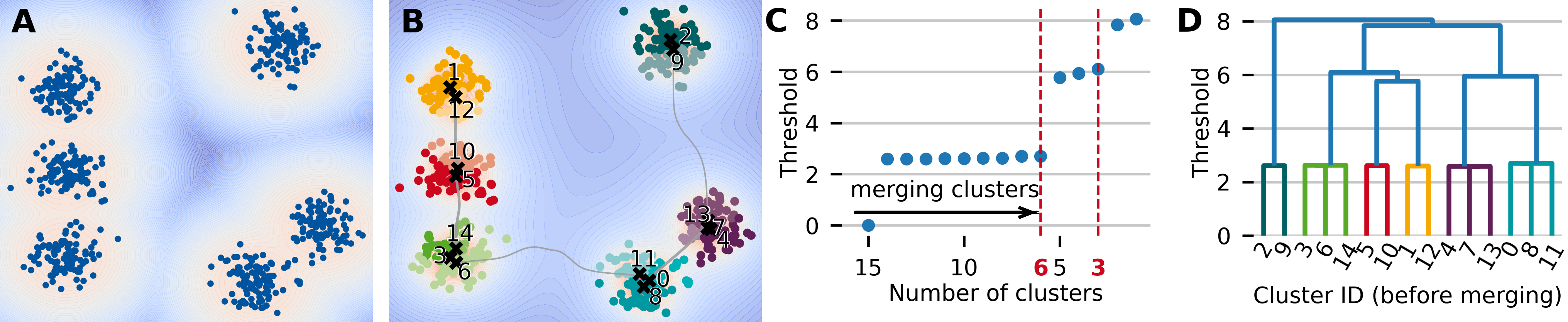}
    \caption{ Overview of t-NEB clustering procedure.
    \textbf{A}:~Illustrative toy dataset with hierarchical density model consisting of six Gaussians. Shading: probability density. Points: samples.
    \textbf{B}:~Overclustering using a Student's $t$ mixture model with 15 components. Centers are marked by `x' and colors indicate overclustered assignments. Lines are maximum density paths, width indicating the minimum density on the path which we use as similarity for cluster merging.
    \textbf{C}:~We iteratively merge clusters starting with the minimum density as threshold. Merges are
    solely based on the initial overclustered partition, resuling in consistent clusterings at any level of granularity.
    At both three and six the threshold jumps, indicating meaningful clusterings.
    \textbf{D}:~Dendrogram of the hierarchical merging procedure.
    The thresholds from \textbf{C} leading to three or six clusters are clearly visible, showing that the algorithm has taken up on the hierarchical nature of the dataset.
    }
    \label{fig:intro}
\end{figure}

A common solution to this challenge is to first overcluster the data, then merge clusters using a distance metric or a statistical test
\cite{vardakas2023UniForCE, guan2022smmp, ward1963hierarchical, Weis2022}.
Overcluster-then-merge methods commonly use one algorithm to overcluster the data and an independent strategy to merge those clusters.
We seek a more principled, probabilistically grounded approach and hypothesize that 
basing the merging strategy on a parametric density model is beneficial.

Existing hierarchical methods \citep{vardakas2023UniForCE,guan2022smmp,ward1963hierarchical} often struggle with non-Gaussian high-dimensional data, especially when clear density gaps are absent or class distributions are imbalanced, both of which are common in real-world data.
Only some of those algorithms such as  UniForCE \citep{vardakas2023UniForCE} or HDBSCAN \citep{campello2013density} are able to capture clusters of arbitrary shapes. However, they are mostly density-based and thus prone to merge `lighly touching' clusters due to the way they estimate density.
Moreover, many current hierarchical clustering methods use distance-based overclustering or merging based on dip-statistics \citep{Weis2022, leiber2021dip}, both of which degrade in performance as the dimensionality of the data increases. 
Although dimensionality reduction techniques such as t-SNE \citep{van2008visualizing} or UMAP~\citep{mcinnes2018umap} are sometimes used prior to clustering, preserving inter-point distances in lower dimensions is inherently challenging and, in general, not guaranteed (see Johnson-Lindenstrauss lemma connecting number of datapoints, target dimension, and distortion of distances).
Thus, interpreting clustering on dimensionality-reduced data is conceptually problematic.

We propose a paradigm that focuses on hierarchical structures in high-dimensional data and is able to handle both arbitrary shapes and the lack of clear density gaps, while avoiding dimensionality reduction as a necessary pre-processing step.
It further uses a single density model for both initial clustering and merging strategy, enabling a probabilistically grounded hierarchical clustering  (\cref{fig:intro}). 
Overall, we make the following contributions:

\begin{itemize}[noitemsep, topsep=-5pt, leftmargin=*, itemsep=0pt, partopsep=5pt]
    \item We introduce t-NEB, a hierarchical clustering algorithm  
    that derives a clustering hierarchy directly from a data-driven parametric density model (\cref{sec:methods}).
    \item t-NEB outperforms other hierarchical clusterings 
    on realistic high-dimensional datasets (\cref{sec:HDresults}).%
    \item t-NEB successfully uncovers hierarchies in real-world data (\cref{sec:realdata}).
\end{itemize}

\section{Related Work}

We first discuss methods that adopt a similar overcluster-then-merge procedure. While effective, these approaches are not inherently hierarchical. We then briefly review relevant top-down hierarchical methods before focusing on bottom-up approaches most closely related to our own.

\paragraph{Overcluster-then-merge.}

RCC \citep{shah2017robust}, K-Multiple-Means (KMM) \citep{nie2019k}, and MCKM \citep{li2023multi} formulate clustering as an optimization procedure.
RCC \cite{shah2017robust} starts from singletons and merges points optimizing the regularized Euclidean distance between the points and suggested cluster centers, reducing the amount of suggested cluster centers at every step.
K-Multiple-Means (KMM) \cite{nie2019k} aims to group data points into a specified number of $k$ clusters by modeling the partition as a bipartite graph partitioning problem with a constrained Laplacian rank, optimizing the cluster mean positions.
MCKM \citep{li2023multi} also suggests prototype mean positions and then merges them convexly by adding a sum-of-norms (SON) regularization to control the trade-off between the model error and the number of clusters.
Git \citep{gao2021git} determines a local structure via kernel density estimation, constructs a connectivity graph of local clusters and cuts noisy edges for reading out final clusters using Wasserstein distance.
However, those methods are not hierarchical by design, nor is it straightforward to extract a hierarchy.
Quite similar to Git, PAGA \citep{Wolf2019}, a trajectory inference (TI) method from the cell omics community, determines a local structure via Leiden clustering, constructs a connectivity graph of local clusters and cuts noisy edges for reading out final clusters using statistical tests.
Although PAGA allows analyzing data at different resolutions, it does not enable a hierarchy since the algorithm has to be applied several times with different resolution levels and Leiden clustering do not create a nested structure for different resolutions.
Additionally, PAGA involves dimensionality reduction before detecting local structure, which might change pairwise distances.
Another TI method, PARC \citep{stassen2020parc} creates a $k$-NN graph on the original points, filters the graph and performs community detection on the filtered graph. 
Because of the filtering, it is impossible to extract a hierarchy.
\citet{Weis2022} also designed their method for biological data. 
They overcluster data with Gaussian mixture models, construct a $k$-NN graph ($k=3$) based on fitted centers and use the dip-statistics as edge weights.
Dip-statistics \citep{hartigan1985dip} measures bimodality by comparing the empirical cumulative distribution function (ECDF) to the closest unimodal ECDF.
Another method employing dip-statistics, Dip-DECK \cite{leiber2021dip} performs clustering in the latent space of an autoencoder, where the training is interleaved with cluster merging based on dip-statistics to enhance clusterability. 
It does not create a nested hierarchy though, since points may get re-assigned to a different cluster after autoencoder finetuning.

\paragraph{Top-down hierarchical methods.}
HDBSCAN \citep{campello2013density} is a standard density-based hierarchical algorithm.
It builds upon a minimum spanning tree (MST) using mutual reachability distances -- defined as the maximum of core distances (distance to the $k$-th nearest neighbor) and pairwise point distances. 
Clusters are formed by removing long edges from the MST, yielding a dendrogram that captures varying density levels. 
Although extensions exist \citep{malzer2020hybrid}, the original algorithm remains widely used.
However, density-based hierarchical algorithms are prone to merge `lighly touching' clusters due to the way they estimate density which impacts their performance on real-world data.
Another hierarchical method is Chameleon \citep{karyapis301chameleon}, which constructs a sparse $k$-means graph, partitions it, and dynamically merges the resulting subgraphs. 
While effective, it struggles with singleton clusters and noise \citep{barton2019chameleon}. 
Chameleon 2 addresses these issues via improved graph partitioning and a flood-fill + $k$-NN merging strategy. 
Chameleon 2++ \citep{singh2025chameleon2++} further enhances runtime using approximate $k$-NN. However, neither version provides public code, limiting reproducibility and comparison.

\paragraph{Bottom-up hierarchical methods.}
The most well-known hierarchical clustering method is agglomerative clustering, often used with Ward’s method \citep{ward1963hierarchical}, which merges clusters to minimize the increase in within-cluster variance, in contrast with single-linkage, which merges the closest pair of clusters. 
ROCK \citep{guha2000rock} extends agglomerative clustering to categorical data by using a similarity-based merging criterion.
While agglomerative clustering starts from singleton clusters, several methods begin with prototype subclusters, often obtained by overclustering. 
UniForCE \citep{vardakas2023UniForCE} applies $k$-means++ followed by dip-statistics to decide which clusters to merge. 
SMMP \citep{guan2022smmp} starts with Density Peak Clustering (DPC) \citep{rodriguez2014clustering}, estimates subcluster similarity via a border-link method, and merges them using single-linkage agglomeration. 
Similarly, \citet{mehmood2017clustering} also build on DPC and merge subclusters that share density regions.
Skeleton clustering \citep{wei2022skeleton} and K-mH \citep{peterson2018merging} both overcluster using $k$-means and then merge with agglomerative clustering. 
Skeleton clustering assigns inter-cluster weights based on Voronoi, Face, or Tube densities.
K-mH computes distances based on the probability to misassign a point to an incorrect cluster under a non-homogeneous spherical Gaussian model.
Our method is also bottom-up hierarchical but additionally couples the merging strategy with the clustering through a shared density model which grounds our method probabilistically.
\section{Our method: t-NEB hierarchical clustering}
\label{sec:methods}
\begin{wrapfigure}[7]{r}{0.5\linewidth}
    \centering
    \vspace{-12pt}
    \begin{subfigure}[b]{0.33\linewidth}
        \includegraphics[width=\textwidth]{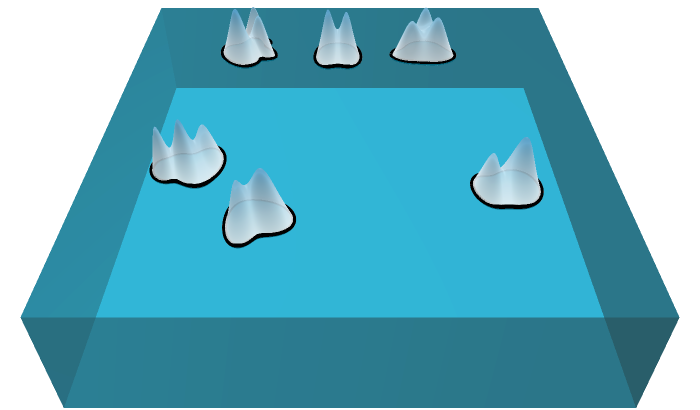}
    \end{subfigure}%
    \begin{subfigure}[b]{0.33\linewidth}
        \includegraphics[width=\textwidth]{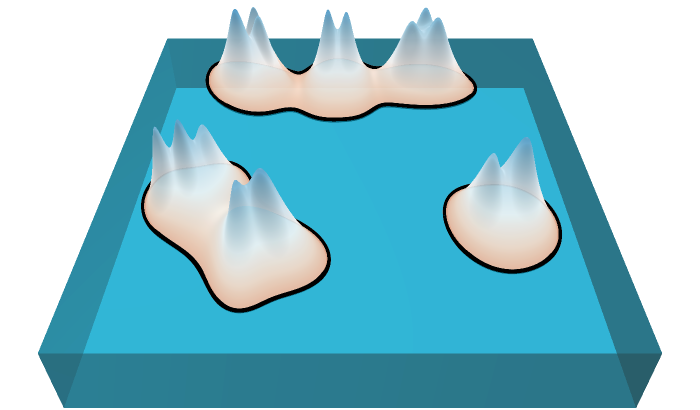}
    \end{subfigure}%
    \begin{subfigure}[b]{0.33\linewidth}
        \includegraphics[width=\textwidth]{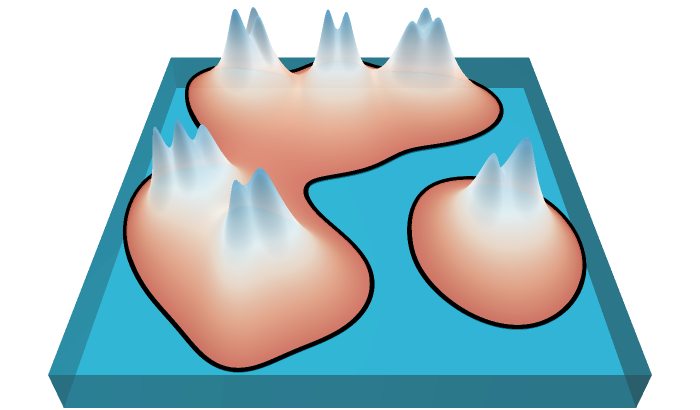}
    \end{subfigure}
    \caption{Density landscape thresholded at different ``water levels'' leading to 6, 3, and 2 clusters.}
    \label{fig:energy-landscapes}
\end{wrapfigure}
Our approach models the underlying density landscape using parametric density estimation and merges modes based on the depth of the valleys separating them -- analogous to islands joining as water levels decrease (\cref{fig:energy-landscapes}).
We define the distance between two clusters as the lowest value along the highest `valley' path between the peaks (the \emph{maximum-density path}).
To operationalize this intuition we introduce t-NEB, a principled probabilistic hierarchical clustering algorithm with three main steps:
\begin{enumerate}[noitemsep, leftmargin=*, itemsep=0pt]
    \item \textit{Density estimation:} 
    We estimate a fine-grained and accurate density model of the data by overclustering with a Student's $t$ mixture model.
    \item \textit{Distance computation:} 
    We merge clusters in dense regions first. Our distance is the minimum along the maximum-density path between clusters, found using nudged elastic band (NEB) optimization. 
    \item \textit{Hierarchy construction:} 
    We build the hierarchy by iteratively merging clusters with the smallest distances. 
    Unlike agglomerative clustering, distances remain fixed after merging.
\end{enumerate}

Below we formally describe the three steps.
The complete t-NEB algorithm is summarized in \cref{alg:t-neb}.

\begin{algorithm2e}[b]
\caption{t-NEB Hierarchical Clustering}
\label{alg:t-neb}
\KwData{data $\mathcal{X}$, number of subclusters $m$}
\KwResult{raw clustering and hierarchy}
\SetCommentSty{mycommfontHighlight}
\tcp{1. STEP: ESTIMATING DENSITY LANDSCAPE BY OVERCLUSTERING}
\SetCommentSty{mycommfont}
tmm $\gets$ TMM.fit$(\mathcal X, m)$ \tcp{density estimation}

$\mathfrak C^+ \gets$ tmm.predict($\mathcal X$) \tcp{initial overclustered partition}

\SetCommentSty{mycommfontHighlight}
\tcp{2. STEP: DISTANCE COMPUTATION}
\SetCommentSty{mycommfont}

$A \gets$  infinity$(m, m)$ \tcp{initialize adjacency matrix}

\For{$i$ \emph{\textbf{in}} range($m$)} {

    \For(\tcp*[h]{focus on nearby clusters in Euclidean space}){$j$ \emph{\textbf{in}} range($m$)} {

        \If{$j$ \emph{\textbf{in}} $\text{kNN}(i)$} {
            $A_{ij} \gets$ compute\_distance\_with\_NEB(tmm, $i$, $j$) \tcp{\cref{alg:nebDistance}}
        }
    }
}

\tcp{if $i$ and $j$ are not closest neighbors, the path goes through other clusters, so we take the minimum density along the minimum spanning tree path}
graph $\gets$ minimum\_spanning\_tree($A$)

\For{$i$ \emph{\textbf{in}} range($m$)} {

    \For{$j$ \emph{\textbf{in}} range($m$)} {
        $A_{ij}\gets$ min\_density\_on\_shortest\_path(graph, $i$, $j$) 
    }
}
\SetCommentSty{mycommfontHighlight}
\tcp{3. STEP: HIERARCHY CONSTRUCTION}
\SetCommentSty{mycommfont}

$H\gets$ hierarchy tree based on $A$ and $\mathfrak C^+$

\Return $H$ %
\end{algorithm2e} %

\paragraph{Preliminaries: Hierarchical clustering.}
In clustering, we aim to partition a dataset \( \mathcal{X} = \{ \mathbf{x}_1, \mathbf{x}_2, \dots, \mathbf{x}_N \} \subset \mathbb{R}^d \) into a clustering $\mathfrak C$ of \( m \) disjoint clusters \( \mathcal{C}_1, \dots, \mathcal{C}_m \)
such that clusters are high-density regions that are separated by low density regions.
Since the optimal clustering partition $\mathfrak C^*$ and the optimal number of clusters $m^*$ are generally unknown in real-world data, a natural choice is using hierarchical clustering which consists of a series of `consistent' clusterings $\mathfrak C_1,\dots,\mathfrak C_m$ consisting of 1 to $m$ clusters, respectively.
Consistent means that for clusterings $\mathfrak C_i,\mathfrak C_{i+1}$ there is exactly one $C_j\in\mathfrak C_i$ which is split into two clusters in $\mathfrak C_{i+1}$ while all other clusters stay the same.
This defines a dedrogram.

\paragraph{1. Density Estimation.}
We initially calculate an overclustering partition $\mathfrak C^+$ comprised of $m$ clusters ($m \gg m^*$), using a Student's $t$ mixture model (TMM).
The Student's $t$ distribution $p_\textit{St}(\mathbf x \mid \mathbf{\mu},\mathbf{\Lambda},\nu)$ generalizes the Gaussian distribution by adjusting the heaviness of tails with an additional degrees of freedom parameter $\nu$ in addition to the means $\mathbf \mu$ and scale matrices~$\mathbf \Lambda$.
The weighted combination of multiple such distributions then defines the density landscape 
$p_\text{MM}(\mathbf{x} \mid \Theta_1,\dots,\Theta_K) =  
\sum_{i=1}^K w_i p_\textit{St}(\textbf x \mid \Theta_i) $  
with distribution parameters $\Theta_i = (\mu_i,\Lambda_i,\nu_i, w_i)$ where
$w_i$ indicates the (normalized) weight of each component.
During the fitting process, we enforce a minimum cluster size and ensure well-behaved clusters by dropping overfitted components as described in \cref{app:hyperparameters}.
This is only necessary for noisy datasets, for which the EM algorithm returns a large number of singleton clusters.
While we could use any energy- or likelihood-based model such as the widely used Gaussian mixture model (GMM), we use TMMs as they provide better density estimates especially on noisy datasets (see \cref{app:extendedResultsTMMvsGMM}).

\begin{figure}
    \begin{minipage}{0.59\linewidth}
        \vspace{0mm}
        \includegraphics[width=\linewidth]{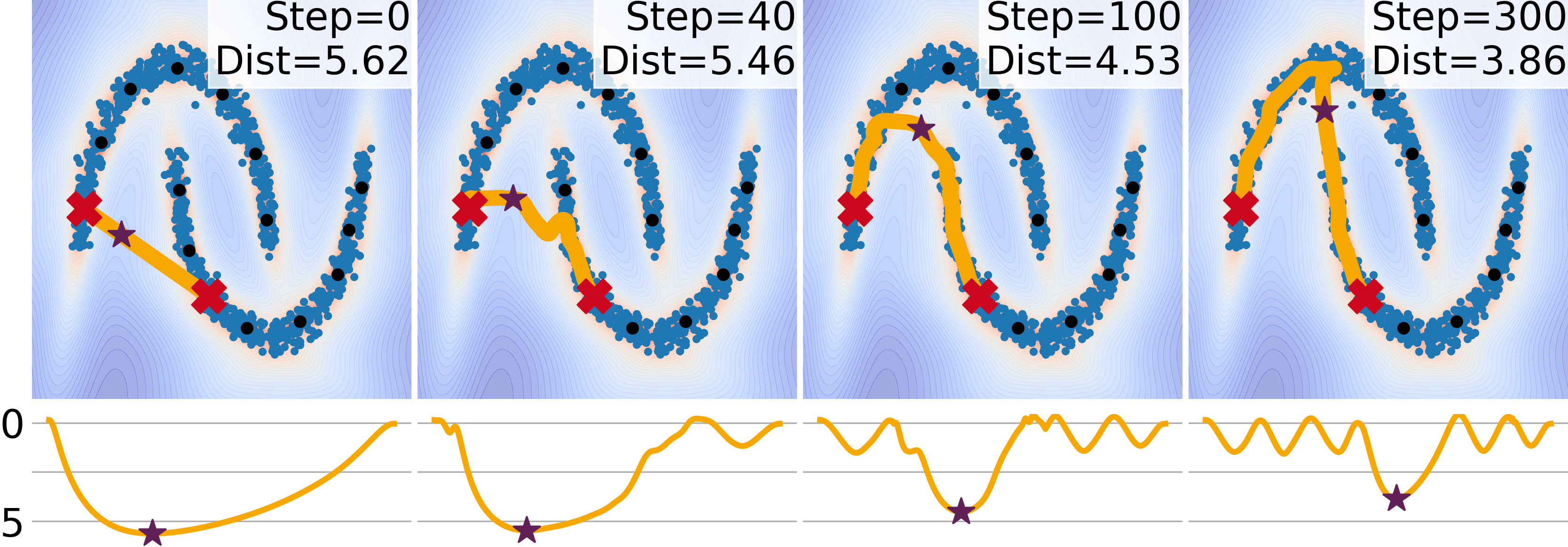}
    \end{minipage}\hfill
    \begin{minipage}{0.4\linewidth}
        \caption{
       Optimization of a maximum-density path (yellow line) using the NEB algorithm. 
        The minimum density along this path is our measure of distance between two mixture components. Bottom: Probability density along the NEB path, estimated from the mixture model.
        }
        \label{fig:optimization_illustration}
    \end{minipage}
\end{figure}

\paragraph{2. Distance computation.}
For each pair of cluster centers, we seek the path of maximum density between the two in the density landscape given by $p_\text{MM}(\mathbf{x} \mid \Theta_1,\dots,\Theta_K)$.
To find this path, we use the \emph{nudged elastic band} (NEB) method \citep{jonsson1998nudged}, a well-established optimization technique in physics which was originally used for finding minimum energy paths between two points on a function landscape. 
NEB samples points along a path between two centers (initially a straight line, as shown in \cref{fig:optimization_illustration}, Step 0; \cref{alg:nebDistance}, Line~1).
It then optimizes their positions to move towards high-density regions using gradient descent (\cref{fig:optimization_illustration} Steps, 40--300; \cref{alg:nebDistance}, Lines~2--5) and returns the lowest density $D(c_1, c_2)$ along the NEB path  as described in \cref{alg:nebDistance}, Line 6.
In order to find the NEB path, we optimize the negative log-likelihood $-\ln p_\text{MM}(\textbf{x} \mid \Theta_1,\dots\Theta_m)$ of the density such that the path is pushed towards more dense regions via gradient descent.
After each gradient descent update, the points in the path $p$ may be no longer equally spaced as each point was moved independently.
We thus resample $l$ points at equidistant intervals before the next gradient descent update for which we interpret $p$ as piecewise-linear and measure distances along the path (Line 5).

\begin{algorithm2e}[t]
\caption{Nudged Elastic Band (NEB) Distance}
\label{alg:nebDistance}
\KwData{Density landscape given by TMM with parameters $\Theta_1,\dots,\Theta_m$, cluster centers $c_1$, $c_2$, steps~$s$, learning rate $\eta$, points per path $\ell$}
\KwResult{Distance between cluster centers $c_1$ and $c_2$}

\tcp{sample a path $\mathbf{p}=p_0,\dots,p_\ell$ of $\ell$ equidistant points on the line from $c_1$ to $c_2$\!\!}

$p_i = c_1 + \frac{i}{\ell}(c_2-c_1), \quad i = 0,1, \dots, \ell $

\For{s \emph{\textbf{steps}} }{
    $\mathcal{L}(\mathbf{p}) = - \sum_{i=0}^{\ell}\ln p_\text{MM}({p}_i \mid \Theta_1,\dots,\Theta_m)$ \tcp{$-$log-likelihood loss of path $\mathbf{p}$\!\!\!\!}
    \tcp{Shift points toward higher-probability regions via gradient descent}
    $\mathbf{p} = \mathbf{p} - \eta \cdot \nabla_{\mathbf{p}} \mathcal{L}(\mathbf{p})$ 

    $\mathbf{p} =\text{LinearInterpolate}\left( \left\{ ({p}_i, {p}_{i+1}) \right\}_{i=0}^{n-1}, \, \ell \right)$%
    \tcp{resample along updated path}

}
\smallskip
\tcp{return lowest density along the optimized NEB path}
\Return $D(c_1, c_2) = \max_{i} \big( -\ln p_\text{MM}({p}_i\mid\Theta_1,\dots,\Theta_m) \big)$ 
\end{algorithm2e}

Computing NEB paths for all pairs of clusters scales quadratically with the number of clusters, which is problematic when this number is large (e.g., $m>100$). 
Moreover, if clusters $c_1$ and $c_2$ are not close neighbors, the path will likely go through other clusters.
To reduce the computational cost and to avoid local optima during NEB optimization, we compute NEB paths only locally for the ten nearest neighbors in Euclidean distance (see ablation in \cref{app:hyperparameters}).
We derive the remaining distances by taking the minimum density along the minimum spanning tree (MST).

\paragraph{3. Hierarchy construction and concrete clusterings.}
We construct a clustering hierarchy by parsing the MST (\cref{fig:intro} \textbf{D}). 
The finest level $\mathfrak C_m$ is given by the overclustered partition \( \mathfrak C^+ \) with \( m \) clusters. 
Starting with the lowest distance $d_\text{min}$, we iteratively merge cluster pairs, resulting in the clusterings $\mathfrak C_{m-1},\dots,\mathfrak C_1$, together with the distances as thresholds defining the hierarchy $H$.
In this way, we transition smoothly from the overclustered partition (tiny islands) to a coarser clustering. 
The NEB paths make sure that we indeed merge the clusters in the order defined by the density landscape~(\cref{fig:energy-landscapes}).
To get a concrete clustering we merge every cluster below a threshold, 
alternatively, we use a target number of clusters to find the cut-off threshold.
Large jumps in the recorded merge thresholds indicate meaningful clusterings (see \cref{fig:intro} \textbf{C} and \cref{fig:interpretability} \textbf{D}). 

Importantly, we do not recompute cluster centers or reassign points during this process -- merges are solely based on the initial overclustered partition. 
This allows us to retrieve a consistent clustering at any level of granularity, a core property of hierarchical clustering.
Since t-NEB merges clusters, the resulting clusters may be of arbitrary shape and are no longer restricted to the ellipsoids of the underlying TMM clustering.
A key aspect of this algorithm is that density estimation, which we do using overclustering with a mixture model, defines both the clusters as well as the order in which they are merged (see the `water levels' intuition).
The distance calculation using NEB is only needed to extract this information from the density model.
This contrasts with most other overcluster-and-merge algorithms where overclustering and merging are effectively independent, using e.g. $k$-means for overclustering and dip-statistics for merging.

\paragraph{Complexity analysis.}
The fitting of the mixture model with $m$ components is linear in the number of data points $N$ (for a fixed number of EM steps) and components, thus, taking $O(N m)$ time.
The complexity of computing the distances with NEB depends on the number of steps $s$ to optimize the path between the centers, as well as the number of points $\ell$ on that path. %
Similar to the number of EM steps, we assume $s$ and $\ell$ to be well-chosen constants.
Since we need to execute the distance computation for every cluster center
to each of their $k$ nearest neighbors, this step is in $O(m k)$ time. %
The creation of the MST is linear in the number of optimized NEB paths between cluster centers $O(m  k)$.
Computing the $m^2$ pairwise distances between cluster centers is in $O(m^2)$ since computing each distance is possible in amortized constant time.
The overall complexity of our hierarchical clustering algorithm results in $O(Nm+m  k+m^2)= O(Nm+m^2)$.
\section{Experimental setup}
\label{sec:experiments}

\paragraph{Datasets.}
We want to mimic a realistic scenario and thus evaluate on noisy data without clear density gaps.
We create density-connected data using the Densired generator \citep{DENSIRED} that produces high-dimensional datasets by constructing a skeleton through a random walk and sampling points from distributions centered on this skeleton.
In our experiments, points are either confined 
within hyperspheres with a hard boundary (Densired `circles') or follow a Student's $t$-distribution (Densired `Stud-t').
For all datasets, the classes are touching, meaning that the average within-cluster distance is larger than the minimum distance to another cluster. 
While the `circles' datasets are linearly separable, on the \mbox{`Stud-t'} datasets a simple multi-layer perceptron achieves only around 95\% accuracy.
More details on the generation procedure and properties of the datasets can be found in \cref{app:dataset_gen}.
We further evaluate our method on machine learning embeddings provided by the MNIST-Nd dataset \citet{turishcheva2024mnist}\footnote{Vector embeddings are an important part of modern ML-based data analysis workflows in science. Hence, using embeddings generated by neural nets as a test dataset is highly relevant in practice.}. 
Those embeddings are mixture-VAE embeddings of the MNIST dataset and thus a representative of deep neural network embeddings.
We use 8, 16, 32, and 64 dimensional versions of all datasets in our experiments.
We also performed comparisons on `real-world' datasets from \citet{vardakas2023UniForCE} (see \cref{app:uniforcedata}). %
To evaluate our method's ability to derive reasonable hierarchies for real-world data, we use the well-established transcriptomic dataset by \citet{tasic2018shared}.

\paragraph{Baselines.}
We compare our method against recent
hierarchical distance-based algorithms UniForCE \citep{vardakas2023UniForCE} and SMMP \citep{guan2022smmp} as well as agglomerative clustering using Ward's linkage \citep{ward1963hierarchical}.
As an example of algorithms that merge clusters using dip-statistics we use the method from \citet{Weis2022} under the name GWG-dip . 
We also compare against the state-of-the-art non-hierarchical clustering algorithms  HDBSCAN \citep{campello2013density} and Leiden \citep{traag2019louvain}.
Further, we show the performance of pure Gaussian mixture models without cluster merging.
We tune hyperparameters for each baseline individually. 
For t-NEB, we overcluster with 25 mixture components (20 for MNIST-Nd) and otherwise chose hyperparameters that we found to be stable (details and concrete values in \cref{app:hyperparameters}).%

\paragraph{Evaluation.}
We use the adjusted Rand index (ARI) \citep{hubert1985comparing} to compare computed clusterings against the ground truth.
An ARI of 1 indicates perfect agreement (up to label permutations), 0 reflects random partitioning, and negative values suggest worse-than-chance alignment.
Due to most clustering algorithms being stochastic, we report the best out of ten runs while evaluating the degree of stochasticity in \cref{app:stability}.
Other cluster evaluation metrics qualitatively match
(see \cref{app:moremetrics}).

\section{Results}
\label{sec:results}

\begin{figure}[t]
    \centering
    \includegraphics[width=\linewidth]{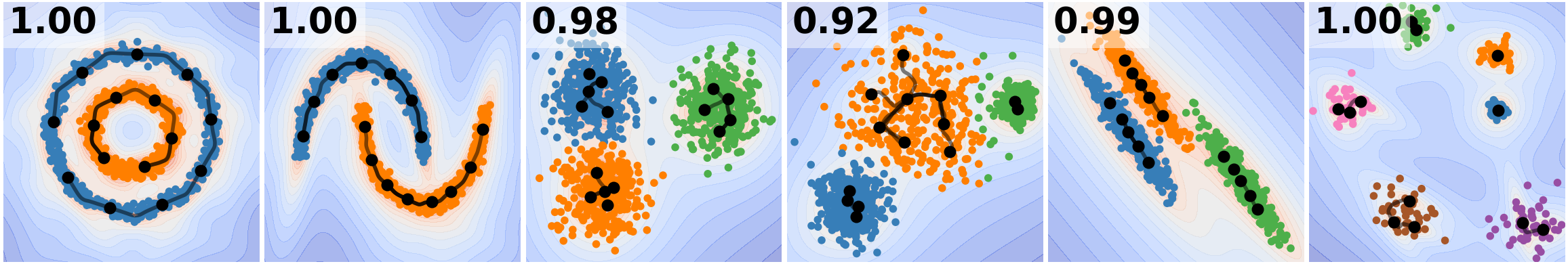}
    \caption{
    t-NEB clustering performance (ARI) on 2D datasets, best of ten runs.
    Background shows the density landscape induced by the mixture model which is the basis for the merging procedure.
    }
    \label{fig:2d-datasets-small}
\end{figure}

We start with a proof-of-concept experiment on standard 2D clustering datasets from \citet{laborde2023sparse} and \citet{scikit-learn} on which t-NEB achieves almost perfect results (\cref{fig:2d-datasets-small}).
A strength of t-NEB is that it is able to solve both the density connected datasets (circles and moons) as well as the almost parallel elongated blobs.
\cref{app:2d} contains further analysis and a comparison with other clustering algorithms.

\paragraph{t-NEB performs competitively on high-dimensional data.}
\label{sec:HDresults}
\begin{table}
\caption{Clustering performance (ARI) on higher-dimensional datasets (best of ten runs). 
t-NEB exhibits state-of-the-art performance across datasets and g-NEB is able to solve the Densired `circles' datasets perfectly.
We highlight the \smash{\colorbox{darkblue!60}{1st}}, \smash{\colorbox{darkblue!30}{2nd}}, and \smash{\colorbox{darkblue!10}{3rd}} best performing algorithms.}
\label{tab:overview_hd}
\resizebox{\textwidth}{!}{
\begin{tabular}{lcccc@{\hspace{8mm}}cccc@{\hspace{8mm}}cccc}
\toprule
 & \multicolumn{4}{c}{Densired `circles'} & \multicolumn{4}{c}{Densired `Stud-t'} & \multicolumn{4}{c}{MNIST-Nd} \\
 & 8D & 16D & 32D & 64D & 8D & 16D & 32D & 64D & 8D & 16D & 32D & 64D \\
\midrule
Agglomerative Clustering\!\!\!\! & 0.68 & 0.66 & 0.59 & 0.75 & 0.56 & {\cellcolor{darkblue!10}} 0.87 & {\cellcolor{darkblue!10}} 0.90 & 0.64 & 0.80 & 0.68 & 0.62 & 0.49 \\
HDBSCAN & 0.00 & 0.00 & 0.44 & 0.00 & 0.01 & 0.00 & 0.00 & 0.00 & 0.03 & 0.06 & 0.07 & 0.07 \\
Gaussian Mixture & 0.79 & 0.77 & 0.69 & 0.91 & 0.74 & 0.68 & 0.49 & 0.42 & {\cellcolor{darkblue!60}} 0.89 & 0.74 & {\cellcolor{darkblue!10}} 0.73 & {\cellcolor{darkblue!10}} 0.62 \\
Leiden & 0.83 & 0.77 & 0.76 & 0.89 & {\cellcolor{darkblue!60}} 0.89 & {\cellcolor{darkblue!30}} 0.93 & {\cellcolor{darkblue!30}} 0.91 & {\cellcolor{darkblue!60}} 0.80 & {\cellcolor{darkblue!60}} 0.89 & {\cellcolor{darkblue!60}} 0.92 & {\cellcolor{darkblue!60}} 0.93 & {\cellcolor{darkblue!30}} 0.71 \\
GWG-dip & {\cellcolor{darkblue!30}} 0.98 & {\cellcolor{darkblue!60}} 1.00 & {\cellcolor{darkblue!60}} 1.00 & {\cellcolor{darkblue!30}} 0.98 & 0.57 & 0.78 & 0.50 & 0.09 & {\cellcolor{darkblue!60}} 0.89 & {\cellcolor{darkblue!10}} 0.78 & 0.44 & 0.56 \\
UniForCE & 0.32 & 0.24 & 0.47 & 0.14 & 0.25 & 0.34 & 0.45 & {\cellcolor{darkblue!10}} 0.78 & 0.13 & 0.13 & 0.05 & 0.03 \\
SMMP & 0.81 & 0.74 & 0.67 & 0.90 & 0.15 & 0.28 & 0.52 & 0.39 & 0.36 & 0.56 & 0.23 & 0.13 \\
\midrule
g-NEB (ours) & {\cellcolor{darkblue!60}} 1.00 & {\cellcolor{darkblue!60}} 1.00 & {\cellcolor{darkblue!60}} 1.00 & {\cellcolor{darkblue!60}} 0.99 & {\cellcolor{darkblue!10}} 0.77 & 0.72 & 0.74 & 0.34 & 0.77 & 0.62 & 0.56 & 0.51 \\
t-NEB (ours) & {\cellcolor{darkblue!10}} 0.92 & {\cellcolor{darkblue!60}} 1.00 & {\cellcolor{darkblue!10}} 0.94 & {\cellcolor{darkblue!10}} 0.96 & {\cellcolor{darkblue!60}} 0.89 & {\cellcolor{darkblue!60}} 0.94 & {\cellcolor{darkblue!60}} 0.94 & {\cellcolor{darkblue!30}} 0.79 & 0.77 & {\cellcolor{darkblue!60}} 0.92 & {\cellcolor{darkblue!30}} 0.79 & {\cellcolor{darkblue!60}} 0.73 \\
\bottomrule
\end{tabular}}
\end{table} On the high-dimensional datasets we observe that t-NEB generally performs among the best methods (\cref{tab:overview_hd}).
While g-NEB excels as the hard-boundary `circles' dataset, t-NEB performs best on the more noisy Densired `Stud t' data and on MNIST-Nd.
Since the only difference between g-NEB and t-NEB is the weight of the tails of the underlying distribution, this indicates that a Student's $t$ distribution is indeed preferable for datasets with more outliers. 
On the latter two datasets, Leiden performed on-par with t-NEB but without constructing a hierarchy. 
GWG-dip is similar to g-NEB but uses a different merging strategy and also performs well on the `circles' dataset and additionally on the low-dimensional MNIST-Nd datasets.
Given that MNIST-Nd was created using a Gaussian mixture prior, the good performance of (pure) GMMs on this dataset is expected and it is even more impressive that from 16D onwards both t-NEB and Leiden are clearly better while overclustering with GMMs worked worse than using TMMs.
The performance of both UniForCE and SMMP is rather low overall.
Exceptions are SMMP on the hard-boundary `circles' dataset and UniForCE on the 64D `Stud t' dataset.
We assume that the lack of clear density gaps is problematic for both algorithms, and that the good performance of UniForCE on the 64D `Stud t' dataset is an outlier as the other runs range from 0.31 to 0.58.
The lack of clear density gaps is presumably also why HDBSCAN completely fails on the datasets.
For a visual illustration of the clustering performance using t-SNE \citep{van2008visualizing} as well as more algorithms see \cref{app:extendedResults}. 
There we also show that t-NEB also excels on UniForCE's smaller datasets with larger density gaps.
Overall, t-NEB achieves state-of-the-art clustering performance while additionally providing a hierarchy.

\begin{figure}
\begin{minipage}{0.5\textwidth}
    
    \includegraphics[width=\textwidth]{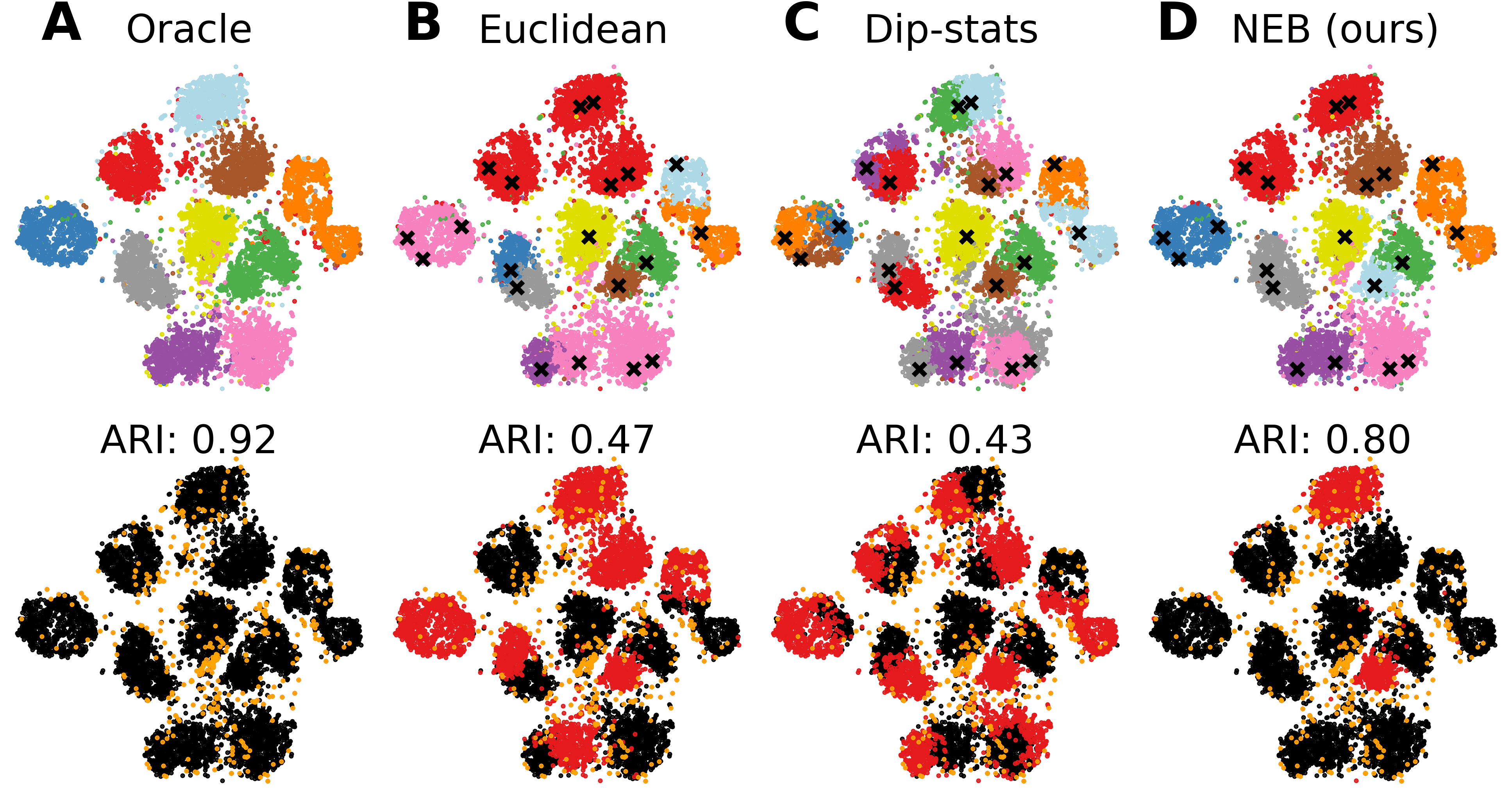}
\end{minipage}%
\begin{minipage}{0.5\textwidth}
    \resizebox{\textwidth}{!}{
\begin{tabular}{p{1.2cm}r|l|lll}
    \toprule
        Dataset & Dim\!\! & Oracle & Euclidean & Dip-stats & NEB (ours) \\ 
        \midrule
        \multirow{4}{1.5cm}{Densired `circles'} & 8  & 0.99 \small{± 0.00} & 0.83 \small{± 0.05} & 0.17 \small{± 0.05} & \textbf{0.92 \small{± 0.02}} \\ 
        & 16 & 1.00 \small{± 0.00} & 0.65 \small{± 0.02} & 0.12 \small{± 0.08} & \textbf{0.96 \small{± 0.04}} \\ 
        & 32 & 1.00 \small{± 0.00} & 0.70 \small{± 0.11} & 0.12 \small{± 0.06} & \textbf{0.89 \small{± 0.06}} \\ 
        & 64 & 1.00 \small{± 0.00} & 0.79 \small{± 0.06} & 0.15 \small{± 0.06} & \textbf{0.94 \small{± 0.02}} \\ \hline
        \multirow{4}{1.5cm}{Densired `Stud-t'} & 8  & 0.88 \small{± 0.00} & \textbf{0.86 \small{± 0.04}} & 0.21 \small{± 0.02} & \textbf{0.85 \small{± 0.05}} \\ 
        & 16 & 0.94 \small{± 0.00} & 0.73 \small{± 0.09} & 0.34 \small{± 0.09} & \textbf{0.85 \small{± 0.00}} \\ 
        & 32 & 0.94 \small{± 0.00} & \textbf{0.87 \small{± 0.02}} & 0.42 \small{± 0.09} & \textbf{0.87 \small{± 0.02}} \\ 
        & 64 & 0.78 \small{± 0.03} & \textbf{0.66 \small{± 0.09}} & 0.54 \small{± 0.14} & \textbf{0.66 \small{± 0.09}} \\ \hline
        \multirow{4}{1.5cm}{MNIST-Nd} & 8  & 0.89 \small{± 0.01} & \textbf{0.77 \small{± 0.03}} & 0.31 \small{± 0.07} & 0.68 \small{± 0.05} \\ 
        & 16 & 0.92 \small{± 0.01} & 0.33 \small{± 0.13} & 0.32 \small{± 0.06} & \textbf{0.78 \small{± 0.04}} \\ 
        & 32 & 0.89 \small{± 0.02} & 0.17 \small{± 0.03} & 0.26 \small{± 0.06} & \textbf{0.76 \small{± 0.06}} \\ 
        & 64 & 0.65 \small{± 0.04} & 0.10 \small{± 0.02} & 0.14 \small{± 0.06} & \textbf{0.55 \small{± 0.06}} \\
        \bottomrule
\end{tabular} %
}
\end{minipage}
    \caption{
    Comparison of merging strategies.
    Merging 25 initial components from a Student's $t$ mixture model to 10 clusters on MNIST-Nd 16D. 
    \textbf{A}: Oracle. Assigns components to classes based on highest overlap with ground-truth labels.
    \textbf{B}: Euclidean distance. 
    \textbf{C}: Dip statistic.
    \textbf{D}: NEB.
    \emph{Top}: Clustering results. 
    \emph{Bottom}: Error plot. 
    Black: Correct assignments. 
    Orange: Unavoidable errors (due to underlying mixture components containing points from several classes).
    Red: Misassigned points due to merging errors.
    \textbf{Table}: Mean ARI\,($\pm$\,std) over 10 seeds of the presented merging strategies. %
    }
    \label{fig:join_strategies}
\end{figure}

\paragraph{NEB outperforms common merging strategies.}
t-NEB consists of two main components: the density model and the merging strategy.
We analyze merging strategies by overclustering high-dimensional datasets using a TMM with 25 components and comparing four merging strategies: an oracle, the dip-statistic, Euclidean distance, and NEB. 
The oracle uses label information to assign each component to the cluster it overlaps with most (\cref{fig:join_strategies}A, top), providing an upper bound limited by misassignments from the mixture model (\cref{fig:join_strategies}A, bottom, orange points). 
For example, on the 16D MNIST-Nd dataset, it achieves an ARI of 0.92 instead of a perfect alignment as some mixture components contained points from more than one class, causing unavoidable errors.

For the three other strategies (dip statistic, Euclidean distance, NEB), clusters are merged iteratively to match the target number. 
Dip-statistic and Euclidean distance only partially recover the cluster structure, with ARI values of 0.43 and 0.47, respectively (\cref{fig:join_strategies}B--C). 
In contrast, NEB merges clusters more accurately, making only one mistake, merging the red and light blue cluster (digits 7 and 9) before the two components of the green cluster (digit 2), resulting in a significantly higher ARI of 0.80 (\cref{fig:join_strategies}D).
This pattern holds across all datasets and dimensionalities (Table in \cref{fig:join_strategies}) -- note that the plots A--D show a concrete example while the table shows averages such that the numbers differ while the overall picture remains the same.

\paragraph{t-NEB derives meaningful hierarchies: MNIST-Nd case study.}
\begin{figure}[b!]
    \centering
\includegraphics[width=\linewidth]{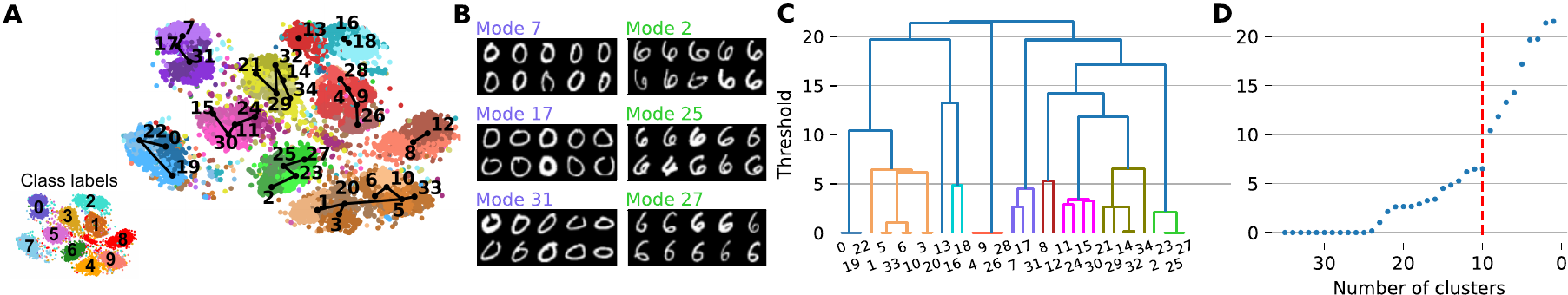}
    \caption{
    \textbf{A}: MNIST-Nd 32D overclustered with 35 initial components and merged to 10 clusters using t-NEB. 
    Each major color is one of the ten clusters, components are indicated by different color shades. %
    Numbers are component IDs. %
    \textbf{B}: MNIST pictures belonging to different components from two clusters.
    \textbf{C}: Corresponding dendrogram.
    \textbf{D}: Threshold to achieve a specific number of clusters. %
    }
    \label{fig:interpretability}
\end{figure}
Our method introduces a hierarchical structure for exploratory data analysis, helping to reveal patterns in complex datasets.
In this section we present a case study on the MNIST-Nd dataset to validate the interpretability of t-NEB.
While the high-level structure of ten digit classes is well established, we investigate whether our method detects these classes and whether the overclustered substructure is meaningful. 
For example, the purple cluster in the top-left corner of \cref{fig:interpretability}A contains overclustered components 7, 17, and 31. 
Examining the original images, we find that digits in component 31 are narrower and more tilted than those in components 7 and 17 (\cref{fig:interpretability}B). 
This relationship is reflected in the dendrogram, where components 7 and 17 merge first, followed by component 31, suggesting that component 31 is slightly more distinct (\cref{fig:interpretability}C).
A similar pattern appears in the green cluster located at the bottom middle of the t-SNE map, containing components 2, 23, 25, and 27. 
The original images show that digits in component 2 tend to be left-tilted, whereas components 25 and 27 are more upright (\cref{fig:interpretability}B). 
The dendrogram confirms this structure, as components 25 and 27 merge earlier, reflecting their similarity. 
The interpretability of the distance is further supported by the minimal threshold needed to achieve a specific number of clusters after merging (\cref{fig:interpretability}D):
a notable increase in the threshold occurs when the number of clusters drops below 10, suggesting a major structural change in the data organization.

\begin{figure}[]
    \centering
\includegraphics[width=\linewidth]{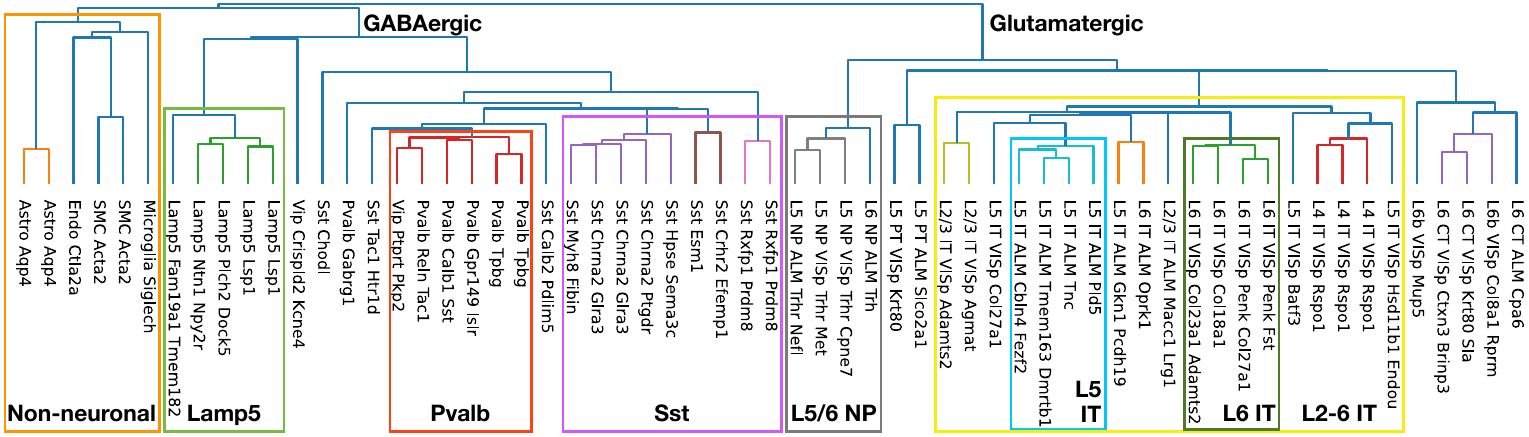}
    \caption{
    t-NEB hierarchy on transcriptomic dataset by \citet{tasic2018shared}.
    Early splits identify three main groups: non-neuronal (orange, left), GABAergic (middle) and Glutamatergic cell types (right).
    The color-coded groups highlight several more fine-grained accurate cell types. 
    For comparison with the original hierarchy see \cref{fig:tasic-app}.
    }
    \label{fig:realhierarchy}
\end{figure}

\paragraph{t-NEB successfully catches hierarchy of real-world transcriptomic cell-types.}
\label{sec:realdata}
Hierarchical clustering is a natural fit for analyzing cell type hierarchies, particularly because these hierarchies remain uncertain for many cell types and brain areas.
We use a well-established transcriptomic dataset from \citet{tasic2018shared} which contains 20,502 cells and 75 (fine-grained) cell types after preprocessing (details in \cref{app:realhierarchy}). 
We apply t-NEB to the 50-dimensional data. 
t-NEB successfully matches the hierarchy from the original paper (our hierarchy in \cref{fig:realhierarchy}, original hierarchy in \cref{app:realhierarchy}).
In more detail, t-NEB accurately separates non-neuronal from neuronal cells, as well as neuronal cells into GABAergic and Glutamatergic cell types, reflecting their distinctiveness (\cref{fig:realhierarchy} non-neuronal: left, orange; GABAergic: middle; Glutamatergic: right).
Zooming in at the GABAergic cells, t-NEB correctly aggregates \emph{Lamp5} cells (green) and merges subgroups appropriately, \emph{`Lamp5 Fam19a1 Tmem182'} are grouped together with \emph{`Lamp5 Ntn1 Npy2r'} and \emph{`Lamp5 Plch2 Dock5'} before the subgroups merge. 
\emph{Pvalb} (red) and \emph{Sst} (pink) cells cluster closely together in the hierarchy and are joined before combining with \emph{Vip}, \emph{Sncg}, and others, again matching the original paper. %
When analyzing the hierarchy for the Glutamatergic cell types (right half of the plot starting with gray), we see that t-NEB captures the exact local match with \citet{tasic2018shared} for \emph{L5/6 NP} cells (gray).
Further, all IT cells (yellow) are accurately merged across layers (L2/3 to L6), and light blue and dark green reflect a correct local hierarchy within the IT cells.

\section{Discussion}
We presented t-NEB, a probabilistically grounded hierarchical clustering method where both clustering and hierarchy computation is based on a density model.
t-NEB achieves state-of-the-art performance on various datasets.
Further, case studies on MNIST-Nd and \citet{tasic2018shared} show that the computed hierarchy uncovers meaningful fine-grained structure.

\textbf{Limitations.} 
Our method's main limitation is its reliance on a mixture model for density estimation.
Although a parametric density model improves over nearest-neighbor relationships between individual points, estimating density in high-dimensional spaces remains a challenge. 
While tested up to 64 dimensions, scaling to thousands may face computational and qualitative issues, requiring significantly more points for adequate density estimation. 
The underlying mixture model also makes our method dependent on random initializations.
Although our ablations show that t-NEB is robust against initialization (\cref{app:stability}), a fully deterministic approach would be preferable. 
However, the merging procedure is deterministic (up to numerics), as it uses fixed initial conditions and performs gradient descent along the `path' between cluster pairs without sampling.
Furthermore, gains may be possible with alternative minimum energy path methods proposed in the physics community \citep{sheppard2008optimization}.

\textbf{Future work.} Replacing the underlying density estimation model could be a promising direction to expand our work. 
For example, \citet{izmailov2020semi} use normalizing flows for semi-supervised clustering.
In cases like transcriptomics, more biologically plausible mixture models could be used.
For example, \citet{harris2018classes} fit a mixture of sparse multivariate negative binomial distributions to derive hierarchical cell-type clustering using the Bayesian Information Criterion (BIC). 
Since both BIC \cite{lu2011dependence} and density estimation depend on sample size in different ways, it would be interesting to compare it with our merging procedure.
Moreover, multidimensional versions of dip-statistics recently emerged, e.g. mud-pod \cite{kolyvakis2023multivariate} or the folding test \cite{siffer2018your}. 
Both methods rely on Monte Carlo sampling, so comparing them with t-NEB in terms of accuracy and computational efficiency, especially for high-dimensional data, is also worthwhile.

\section*{Acknowledgements}
We thank Kenneth Harris, Ayush Paliwal, Philipp Berens, Dmitry Kobak, Sebastian Damrich, Anna Beer, and Pascal Weber for insightful discussions.
We thank Rasmus Steinkamp for the technical support. %

We gratefully acknowledge the computing time granted by the Resource Allocation Board and provided on the supercomputer Emmy/Grete at NHR-Nord@Göttingen as part of the NHR infrastructure. The calculations for this research were conducted with computing resources under the project nim00012.

This project has received funding from the European Research Council (ERC) under the European Union’s Horizon Europe research and innovation programme (Grant agreement No. 101041669)

\FloatBarrier
\newpage
{
\small
\bibliographystyle{unsrtnat}
\bibliography{references}

\begin{thebibliography}{54}
\providecommand{\natexlab}[1]{#1}
\providecommand{\url}[1]{\texttt{#1}}
\expandafter\ifx\csname urlstyle\endcsname\relax
  \providecommand{\doi}[1]{doi: #1}\else
  \providecommand{\doi}{doi: \begingroup \urlstyle{rm}\Url}\fi

\bibitem[Gagolewski et~al.(2021)Gagolewski, Bartoszuk, and
  Cena]{gagolewski2021cluster}
Marek Gagolewski, Maciej Bartoszuk, and Anna Cena.
\newblock Are cluster validity measures (in) valid?
\newblock \emph{Information Sciences}, 581:\penalty0 620--636, 2021.

\bibitem[Toma{\v{s}}ev and Radovanovi{\'c}(2016)]{tomavsev2016clustering}
Nenad Toma{\v{s}}ev and Milo{\v{s}} Radovanovi{\'c}.
\newblock Clustering evaluation in high-dimensional data.
\newblock In \emph{Unsupervised learning algorithms}, pages 71--107. Springer,
  2016.

\bibitem[Harris et~al.(2018)Harris, Hochgerner, Skene, Magno, Katona,
  Bengtsson~Gonzales, Somogyi, Kessaris, Linnarsson, and
  Hjerling-Leffler]{harris2018classes}
Kenneth~D Harris, Hannah Hochgerner, Nathan~G Skene, Lorenza Magno, Linda
  Katona, Carolina Bengtsson~Gonzales, Peter Somogyi, Nicoletta Kessaris, Sten
  Linnarsson, and Jens Hjerling-Leffler.
\newblock Classes and continua of hippocampal ca1 inhibitory neurons revealed
  by single-cell transcriptomics.
\newblock \emph{PLoS biology}, 16\penalty0 (6):\penalty0 e2006387, 2018.

\bibitem[Scala et~al.(2021)Scala, Kobak, Bernabucci, Bernaerts, Cadwell,
  Castro, Hartmanis, Jiang, Laturnus, Miranda, et~al.]{scala2021phenotypic}
Federico Scala, Dmitry Kobak, Matteo Bernabucci, Yves Bernaerts,
  Cathryn~Ren{\'e} Cadwell, Jesus~Ramon Castro, Leonard Hartmanis, Xiaolong
  Jiang, Sophie Laturnus, Elanine Miranda, et~al.
\newblock Phenotypic variation of transcriptomic cell types in mouse motor
  cortex.
\newblock \emph{Nature}, 598\penalty0 (7879):\penalty0 144--150, 2021.

\bibitem[Zeng(2022)]{zeng2022cell}
Hongkui Zeng.
\newblock What is a cell type and how to define it?
\newblock \emph{Cell}, 185\penalty0 (15):\penalty0 2739--2755, 2022.

\bibitem[Vardakas et~al.(2023)Vardakas, Kalogeratos, and
  Likas]{vardakas2023UniForCE}
Georgios Vardakas, Argyris Kalogeratos, and Aristidis Likas.
\newblock Uniforce: The unimodality forest method for clustering and estimation
  of the number of clusters.
\newblock \emph{arXiv preprint arXiv:2312.11323}, 2023.

\bibitem[Guan et~al.(2022)Guan, Li, He, Zhu, Chen, and Si]{guan2022smmp}
Junyi Guan, Sheng Li, Xiongxiong He, Jinhui Zhu, Jiajia Chen, and Peng Si.
\newblock Smmp: A stable-membership-based auto-tuning multi-peak clustering
  algorithm.
\newblock \emph{IEEE Transactions on Pattern Analysis and Machine
  Intelligence}, 45\penalty0 (5):\penalty0 6307--6319, 2022.

\bibitem[Ward~Jr(1963)]{ward1963hierarchical}
Joe~H Ward~Jr.
\newblock Hierarchical grouping to optimize an objective function.
\newblock \emph{Journal of the American statistical association}, 58\penalty0
  (301):\penalty0 236--244, 1963.

\bibitem[Weis et~al.(2025)Weis, Papadopoulos, Hansel, Lüddecke, Celii, Fahey,
  Wang, Bae, Bodor, Brittain, Buchanan, Bumbarger, Castro, Collman, da~Costa,
  Dorkenwald, Elabbady, Halageri, Jia, Jordan, Kapner, Kemnitz, Kinn, Lee, Li,
  Lu, Macrina, Mahalingam, Mitchell, Mondal, Mu, Nehoran, Popovych, Reid,
  Schneider-Mizell, Seung, Silversmith, Takeno, Torres, Turner, Wong, Wu, Yin,
  Yu, Reimer, Berens, Tolias, and Ecker]{Weis2022}
M.~Weis, S.~Papadopoulos, L.~Hansel, T.~Lüddecke, B.~Celii, P.~Fahey, E.~Wang,
  J.~Bae, A.~Bodor, D.~Brittain, J.~Buchanan, D.~Bumbarger, M.~Castro,
  F.~Collman, N.~da~Costa, S.~Dorkenwald, L.~Elabbady, A.~Halageri, Z.~Jia,
  C.~Jordan, D.~Kapner, N.~Kemnitz, S.~Kinn, K.~Lee, K.~Li, R.~Lu, T.~Macrina,
  G.~Mahalingam, E.~Mitchell, S.~Mondal, S.~Mu, B.~Nehoran, S.~Popovych,
  R.~Reid, C.~Schneider-Mizell, H.~Seung, W.~Silversmith, M.~Takeno, R.~Torres,
  N.~Turner, W.~Wong, J.~Wu, W.~Yin, S.~Yu, J.~Reimer, P.~Berens, A.~Tolias,
  and A.~Ecker.
\newblock An unsupervised map of excitatory neurons' dendritic morphology in
  the mouse visual cortex.
\newblock \emph{Nature Communications}, 2025.

\bibitem[Campello et~al.(2013)Campello, Moulavi, and
  Sander]{campello2013density}
Ricardo~JGB Campello, Davoud Moulavi, and J{\"o}rg Sander.
\newblock Density-based clustering based on hierarchical density estimates.
\newblock In \emph{Pacific-Asia conference on knowledge discovery and data
  mining}, pages 160--172. Springer, 2013.

\bibitem[Leiber et~al.(2021)Leiber, Bauer, Schelling, B{\"o}hm, and
  Plant]{leiber2021dip}
Collin Leiber, Lena~GM Bauer, Benjamin Schelling, Christian B{\"o}hm, and
  Claudia Plant.
\newblock Dip-based deep embedded clustering with k-estimation.
\newblock In \emph{Proceedings of the 27th ACM SIGKDD Conference on Knowledge
  Discovery \& Data Mining}, pages 903--913, 2021.

\bibitem[Van~der Maaten and Hinton(2008)]{van2008visualizing}
Laurens Van~der Maaten and Geoffrey Hinton.
\newblock Visualizing data using t-sne.
\newblock \emph{Journal of machine learning research}, 9\penalty0 (11), 2008.

\bibitem[McInnes et~al.(2018)McInnes, Healy, and Melville]{mcinnes2018umap}
Leland McInnes, John Healy, and James Melville.
\newblock Umap: Uniform manifold approximation and projection for dimension
  reduction.
\newblock \emph{arXiv preprint arXiv:1802.03426}, 2018.

\bibitem[Shah and Koltun(2017)]{shah2017robust}
Sohil~Atul Shah and Vladlen Koltun.
\newblock Robust continuous clustering.
\newblock \emph{Proceedings of the National Academy of Sciences}, 114\penalty0
  (37):\penalty0 9814--9819, 2017.

\bibitem[Nie et~al.(2019)Nie, Wang, and Li]{nie2019k}
Feiping Nie, Cheng-Long Wang, and Xuelong Li.
\newblock K-multiple-means: A multiple-means clustering method with specified k
  clusters.
\newblock In \emph{Proceedings of the 25th ACM SIGKDD international conference
  on knowledge discovery \& data mining}, pages 959--967, 2019.

\bibitem[Li et~al.(2023)Li, Zhou, Zeng, and Chan]{li2023multi}
Dong Li, Shuisheng Zhou, Tieyong Zeng, and Raymond~H Chan.
\newblock Multi-prototypes convex merging based k-means clustering algorithm.
\newblock \emph{IEEE Transactions on Knowledge and Data Engineering},
  36\penalty0 (11):\penalty0 6653--6666, 2023.

\bibitem[Gao et~al.(2021)Gao, Lin, Tan, Wu, Li, et~al.]{gao2021git}
Zhangyang Gao, Haitao Lin, Cheng Tan, Lirong Wu, Stan Li, et~al.
\newblock Git: Clustering based on graph of intensity topology.
\newblock \emph{arXiv preprint arXiv:2110.01274}, 2021.

\bibitem[Wolf et~al.(2019)Wolf, Hamey, Plass, Solana, Dahlin, G{\"o}ttgens,
  Rajewsky, Simon, and Theis]{Wolf2019}
F~Alexander Wolf, Fiona~K Hamey, Mireya Plass, Jordi Solana, Joakim~S Dahlin,
  Berthold G{\"o}ttgens, Nikolaus Rajewsky, Lukas Simon, and Fabian~J Theis.
\newblock Paga: graph abstraction reconciles clustering with trajectory
  inference through a topology preserving map of single cells.
\newblock \emph{Genome biology}, 20:\penalty0 1--9, 2019.

\bibitem[Stassen et~al.(2020)Stassen, Siu, Lee, Ho, So, and
  Tsia]{stassen2020parc}
Shobana~V Stassen, Dickson~MD Siu, Kelvin~CM Lee, Joshua~WK Ho, Hayden~KH So,
  and Kevin~K Tsia.
\newblock Parc: ultrafast and accurate clustering of phenotypic data of
  millions of single cells.
\newblock \emph{Bioinformatics}, 36\penalty0 (9):\penalty0 2778--2786, 2020.

\bibitem[Hartigan and Hartigan(1985)]{hartigan1985dip}
John~A Hartigan and Pamela~M Hartigan.
\newblock The dip test of unimodality.
\newblock \emph{The annals of Statistics}, pages 70--84, 1985.

\bibitem[Malzer and Baum(2020)]{malzer2020hybrid}
Claudia Malzer and Marcus Baum.
\newblock A hybrid approach to hierarchical density-based cluster selection.
\newblock In \emph{2020 IEEE international conference on multisensor fusion and
  integration for intelligent systems (MFI)}, pages 223--228. IEEE, 2020.

\bibitem[Karyapis et~al.(1999)Karyapis, Han, and Kumar]{karyapis301chameleon}
G~Karyapis, EH~Han, and V~Kumar.
\newblock Chameleon: A hierarchical clustering algorithm using dynamic
  modeling.
\newblock \emph{IEEE Computer, Special Issue on Data Analysis and Mining.
  Efficient Spatial Clustering Algorithm Using Binary Tree}, 301:\penalty0
  881--892, 1999.

\bibitem[Barton et~al.(2019)Barton, Bruna, and Kordik]{barton2019chameleon}
Tomas Barton, Tomas Bruna, and Pavel Kordik.
\newblock Chameleon 2: an improved graph-based clustering algorithm.
\newblock \emph{ACM Transactions on Knowledge Discovery from Data (TKDD)},
  13\penalty0 (1):\penalty0 1--27, 2019.

\bibitem[Singh and Ahuja(2025)]{singh2025chameleon2++}
Priyanshu Singh and Kapil Ahuja.
\newblock Chameleon2++: An efficient chameleon2 clustering with approximate
  nearest neighbors.
\newblock \emph{arXiv preprint arXiv:2501.02612}, 2025.

\bibitem[Guha et~al.(2000)Guha, Rastogi, and Shim]{guha2000rock}
Sudipto Guha, Rajeev Rastogi, and Kyuseok Shim.
\newblock Rock: A robust clustering algorithm for categorical attributes.
\newblock \emph{Information systems}, 25\penalty0 (5):\penalty0 345--366, 2000.

\bibitem[Rodriguez and Laio(2014)]{rodriguez2014clustering}
Alex Rodriguez and Alessandro Laio.
\newblock Clustering by fast search and find of density peaks.
\newblock \emph{science}, 344\penalty0 (6191):\penalty0 1492--1496, 2014.

\bibitem[Mehmood et~al.(2017)Mehmood, El-Ashram, Bie, Dawood, and
  Kos]{mehmood2017clustering}
Rashid Mehmood, Saeed El-Ashram, Rongfang Bie, Hussain Dawood, and Anton Kos.
\newblock Clustering by fast search and merge of local density peaks for gene
  expression microarray data.
\newblock \emph{Scientific reports}, 7\penalty0 (1):\penalty0 45602, 2017.

\bibitem[Wei and Chen(2022)]{wei2022skeleton}
Zeyu Wei and Yen-Chi Chen.
\newblock Skeleton clustering: Graph-based approach for dimension-free
  density-aided clustering.
\newblock In \emph{NeurIPS 2022 Workshop: New Frontiers in Graph Learning},
  2022.

\bibitem[Peterson et~al.(2018)Peterson, Ghosh, and Maitra]{peterson2018merging}
Anna~D Peterson, Arka~P Ghosh, and Ranjan Maitra.
\newblock Merging k-means with hierarchical clustering for identifying
  general-shaped groups.
\newblock \emph{Stat}, 7\penalty0 (1):\penalty0 e172, 2018.

\bibitem[J{\'o}nsson et~al.(1998)J{\'o}nsson, Mills, and
  Jacobsen]{jonsson1998nudged}
Hannes J{\'o}nsson, Greg Mills, and Karsten~W Jacobsen.
\newblock Nudged elastic band method for finding minimum energy paths of
  transitions.
\newblock In \emph{Classical and quantum dynamics in condensed phase
  simulations}, pages 385--404. World Scientific, 1998.

\bibitem[Jahn et~al.(2024)Jahn, Frey, Beer, Leiber, and Seidl]{DENSIRED}
Philipp Jahn, Christian~MM Frey, Anna Beer, Collin Leiber, and Thomas Seidl.
\newblock Data with density-based clusters: A generator for systematic
  evaluation of clustering algorithms.
\newblock In \emph{Joint European Conference on Machine Learning and Knowledge
  Discovery in Databases}, pages 3--21. Springer, 2024.

\bibitem[Turishcheva et~al.(2024)Turishcheva, Hansel, Ritzert, Weis, and
  Ecker]{turishcheva2024mnist}
Polina Turishcheva, Laura Hansel, Martin Ritzert, Marissa~A Weis, and
  Alexander~S Ecker.
\newblock Mnist-nd: a set of naturalistic datasets to benchmark clustering
  across dimensions.
\newblock \emph{arXiv preprint arXiv:2410.16124}, 2024.

\bibitem[Tasic et~al.(2018)Tasic, Yao, Graybuck, Smith, Nguyen, Bertagnolli,
  Goldy, Garren, Economo, Viswanathan, et~al.]{tasic2018shared}
Bosiljka Tasic, Zizhen Yao, Lucas~T Graybuck, Kimberly~A Smith, Thuc~Nghi
  Nguyen, Darren Bertagnolli, Jeff Goldy, Emma Garren, Michael~N Economo,
  Sarada Viswanathan, et~al.
\newblock Shared and distinct transcriptomic cell types across neocortical
  areas.
\newblock \emph{Nature}, 563\penalty0 (7729):\penalty0 72--78, 2018.

\bibitem[Traag et~al.(2019)Traag, Waltman, and Van~Eck]{traag2019louvain}
VA~Traag, L~Waltman, and NJ~Van~Eck.
\newblock From louvain to leiden: guaranteeing well-connected communities. sci.
  rep. 9, 5233, 2019.

\bibitem[Hubert and Arabie(1985)]{hubert1985comparing}
Lawrence Hubert and Phipps Arabie.
\newblock Comparing partitions.
\newblock \emph{Journal of classification}, 2:\penalty0 193--218, 1985.

\bibitem[Laborde et~al.(2023)Laborde, Stewart, Chen, Chen, and
  Brownstein]{laborde2023sparse}
Jose Laborde, Paul~A Stewart, Zhihua Chen, Yian~A Chen, and Naomi~C Brownstein.
\newblock Sparse clusterability: testing for cluster structure in high
  dimensions.
\newblock \emph{BMC bioinformatics}, 24\penalty0 (1):\penalty0 125, 2023.

\bibitem[Pedregosa et~al.(2011)Pedregosa, Varoquaux, Gramfort, Michel, Thirion,
  Grisel, Blondel, Prettenhofer, Weiss, Dubourg, Vanderplas, Passos,
  Cournapeau, Brucher, Perrot, and Duchesnay]{scikit-learn}
F.~Pedregosa, G.~Varoquaux, A.~Gramfort, V.~Michel, B.~Thirion, O.~Grisel,
  M.~Blondel, P.~Prettenhofer, R.~Weiss, V.~Dubourg, J.~Vanderplas, A.~Passos,
  D.~Cournapeau, M.~Brucher, M.~Perrot, and E.~Duchesnay.
\newblock Scikit-learn: Machine learning in {P}ython.
\newblock \emph{Journal of Machine Learning Research}, 12:\penalty0 2825--2830,
  2011.

\bibitem[Sheppard et~al.(2008)Sheppard, Terrell, and
  Henkelman]{sheppard2008optimization}
Daniel Sheppard, Rye Terrell, and Graeme Henkelman.
\newblock Optimization methods for finding minimum energy paths.
\newblock \emph{The Journal of chemical physics}, 128\penalty0 (13), 2008.

\bibitem[Izmailov et~al.(2020)Izmailov, Kirichenko, Finzi, and
  Wilson]{izmailov2020semi}
Pavel Izmailov, Polina Kirichenko, Marc Finzi, and Andrew~Gordon Wilson.
\newblock Semi-supervised learning with normalizing flows.
\newblock In \emph{International conference on machine learning}, pages
  4615--4630. PMLR, 2020.

\bibitem[Lu et~al.(2011)Lu, Ye, and Neuman]{lu2011dependence}
Dan Lu, Ming Ye, and Shlomo~P Neuman.
\newblock Dependence of bayesian model selection criteria and fisher
  information matrix on sample size.
\newblock \emph{Mathematical Geosciences}, 43:\penalty0 971--993, 2011.

\bibitem[Kolyvakis and Likas(2023)]{kolyvakis2023multivariate}
Prodromos Kolyvakis and Aristidis Likas.
\newblock A multivariate unimodality test harnenssing the dip statistic of
  mahalanobis distances over random projections.
\newblock \emph{arXiv preprint arXiv:2311.16614}, 2023.

\bibitem[Siffer et~al.(2018)Siffer, Fouque, Termier, and
  Largou{\"e}t]{siffer2018your}
Alban Siffer, Pierre-Alain Fouque, Alexandre Termier, and Christine
  Largou{\"e}t.
\newblock Are your data gathered?
\newblock In \emph{Proceedings of the 24th acm sigkdd international conference
  on knowledge discovery \& data mining}, pages 2210--2218, 2018.

\bibitem[Dua and Graff(2017)]{Dua:2019}
Dheeru Dua and Casey Graff.
\newblock {UCI} machine learning repository, 2017.
\newblock URL \url{http://archive.ics.uci.edu/ml}.

\bibitem[Cohen et~al.(2017)Cohen, Afshar, Tapson, and
  Van~Schaik]{cohen2017emnist}
Gregory Cohen, Saeed Afshar, Jonathan Tapson, and Andre Van~Schaik.
\newblock {EMNIST}: Extending {MNIST} to handwritten letters.
\newblock In \emph{Int'l Joint Conf. on Neural Networks}, pages 2921--2926.
  IEEE, 2017.

\bibitem[Von~Luxburg(2007)]{von2007tutorial}
Ulrike Von~Luxburg.
\newblock A tutorial on spectral clustering.
\newblock \emph{Statistics and computing}, 17:\penalty0 395--416, 2007.

\bibitem[Frey and Dueck(2007)]{frey2007clustering}
Brendan~J Frey and Delbert Dueck.
\newblock Clustering by passing messages between data points.
\newblock \emph{science}, 315\penalty0 (5814):\penalty0 972--976, 2007.

\bibitem[Comaniciu and Meer(2002)]{comaniciu2002mean}
Dorin Comaniciu and Peter Meer.
\newblock Mean shift: A robust approach toward feature space analysis.
\newblock \emph{IEEE Transactions on pattern analysis and machine
  intelligence}, 24\penalty0 (5):\penalty0 603--619, 2002.

\bibitem[Kobak and Berens(2019)]{kobak2019art}
Dmitry Kobak and Philipp Berens.
\newblock The art of using t-sne for single-cell transcriptomics.
\newblock \emph{Nature communications}, 10\penalty0 (1):\penalty0 5416, 2019.

\bibitem[Strehl and Ghosh(2002)]{strehl2002cluster}
Alexander Strehl and Joydeep Ghosh.
\newblock Cluster ensembles---a knowledge reuse framework for combining
  multiple partitions.
\newblock \emph{Journal of machine learning research}, 3\penalty0
  (Dec):\penalty0 583--617, 2002.

\bibitem[Fowlkes and Mallows(1983)]{fowlkes1983method}
Edward~B Fowlkes and Colin~L Mallows.
\newblock A method for comparing two hierarchical clusterings.
\newblock \emph{Journal of the American statistical association}, 78\penalty0
  (383):\penalty0 553--569, 1983.

\bibitem[Meil{\u{a}}(2007)]{meilua2007comparing}
Marina Meil{\u{a}}.
\newblock Comparing clusterings—an information based distance.
\newblock \emph{Journal of multivariate analysis}, 98\penalty0 (5):\penalty0
  873--895, 2007.

\bibitem[Chan et~al.(2019)Chan, Rao, Huang, and Canny]{chan2019gpu}
David~M Chan, Roshan Rao, Forrest Huang, and John~F Canny.
\newblock Gpu accelerated t-distributed stochastic neighbor embedding.
\newblock \emph{Journal of Parallel and Distributed Computing}, 131:\penalty0
  1--13, 2019.

\bibitem[Wolf et~al.(2018)Wolf, Angerer, and Theis]{wolf2018scanpy}
F~Alexander Wolf, Philipp Angerer, and Fabian~J Theis.
\newblock Scanpy: large-scale single-cell gene expression data analysis.
\newblock \emph{Genome biology}, 19:\penalty0 1--5, 2018.

\bibitem[Bradbury et~al.(2018)Bradbury, Frostig, Hawkins, Johnson, Leary,
  Maclaurin, Necula, Paszke, Vander{P}las, Wanderman-{M}ilne, and
  Zhang]{jax2018github}
James Bradbury, Roy Frostig, Peter Hawkins, Matthew~James Johnson, Chris Leary,
  Dougal Maclaurin, George Necula, Adam Paszke, Jake Vander{P}las, Skye
  Wanderman-{M}ilne, and Qiao Zhang.
\newblock {JAX}: composable transformations of {P}ython+{N}um{P}y programs.
\newblock \emph{GitHub}, 2018.
\newblock URL \url{http://github.com/jax-ml/jax}.

\end{thebibliography}
}

\newpage
\begin{appendix}

\section{Datasets}
\label{app:dataset_gen}
\begin{table}[h]
\begin{center}
\caption{Dataset statistics of the datasets used in our experiments}
\label{tab:datasets}
\begin{tabular}{llrrr}
\toprule
& Dataset Name & \!\!\!\#classes & \#points & dimension \\
\midrule
\multirow{6}{*}{\rotatebox[origin=c]{90}{2D}} &
Noisy circles & 2 & 1000 & 2 \\
& Noisy moons & 2 & 1000 & 2 \\
& Varied density & 3 & 1000 & 2 \\
& Anisotropic blobs & 3 & 1000 & 2 \\
& Gaussian blobs & 3 & 1000 & 2 \\
& \href{#cite.laborde2023sparse}{Clusterlab10} & 6 & 300 & 2 \\
\midrule
\multirow{3}{*}{\rotatebox[origin=c]{90}{8-64D}} &
\href{#cite.DENSIRED}{Densired} `circles' & 6 & 10000 & 8,\,16,\,32,\,64 \\
& \href{#cite.DENSIRED}{Densired} `Student-t' & 6 & 10000 & 8,\,16,\,32,\,64 \\
& \href{#cite.turishcheva2024mnist}{MNIST-Nd} & 10 & 10000 & 8,\,16,\,32,\,64 \\
\midrule
\multirow{7}{*}{\rotatebox[origin=c]{90}{UniForCE}} &
\href{#cite.Dua:2019}{Optdigits}& 10 & 5620 & 8x8 \\
& \href{#cite.Dua:2019}{Pendigits} & 10 & 10992 & 16 \\
& \href{#cite.cohen2017emnist}{EMNIST Balanced Digits (BD)} & 10 & 28000 & 10 \\
& \href{#cite.cohen2017emnist}{EMNIST Balanced Letters (BL)} & 10 & 28000 & 10 \\
& \href{#cite.cohen2017emnist}{EMNIST MNIST (M)} & 10 & 70000 & 10 \\
& \href{#cite.Dua:2019}{Waveform-v1} & 3 & 5000 & 21 \\
& \href{#cite.Dua:2019}{Mice Protein Expression} & 8 & 1080 & 77 \\
\bottomrule
\end{tabular}
\end{center}
\end{table}
 \cref{tab:datasets} summarizes the statistics of the datasets we used in our experiments. 
They vary in number of classes, number of points and data dimensionality.
To create the datasets Densired `circles' we used the original software from \citet{DENSIRED} with the following hyperparameters:
\begin{verbatim}
dim = {8,16,32,64}
radius = 5 
clunum = 6 
core_num = 200 
min_dist = 0.7 
dens_factors = True
step_spread = 0.3 
ratio_con = 0.01
\end{verbatim}

For Densired `Stud-t' we changed \texttt{min\_dist = 1.2} and used the default value of four degrees of freedom for the Student's $t$ distribution.
Because of the heavy tails of this distribution, the clusters not only touch as in the `circles' dataset, but are significantly overlapping.
We verified the latter by training both a linear classifier and a 3-layer MLP which both returned an accuracy of 94-96\% depending on the dataset and random split.

The third group of datasets `UniForCE' from \citet{Dua:2019, cohen2017emnist} are used specifically to compare with two recent clustering algorithms, UniForCE and RCC, which report their performance on these datasets.

\section{Extended clustering results}\label{app:extendedResults}

In addition to the baselines in the main paper, we provide clustering results for MiniBatch $k$-Means, Student-$t$ mixture, Spectral clustering \citep{von2007tutorial}, affinity propagation \citep{frey2007clustering}, MeanShift \citep{comaniciu2002mean}, and PAGA~\citep{Wolf2019}. 

\paragraph{Clustering Performance on 2D Data.}
\label{app:2d}

\begin{figure*}[t!]
    \centering
    \includegraphics[width=0.6\textwidth]{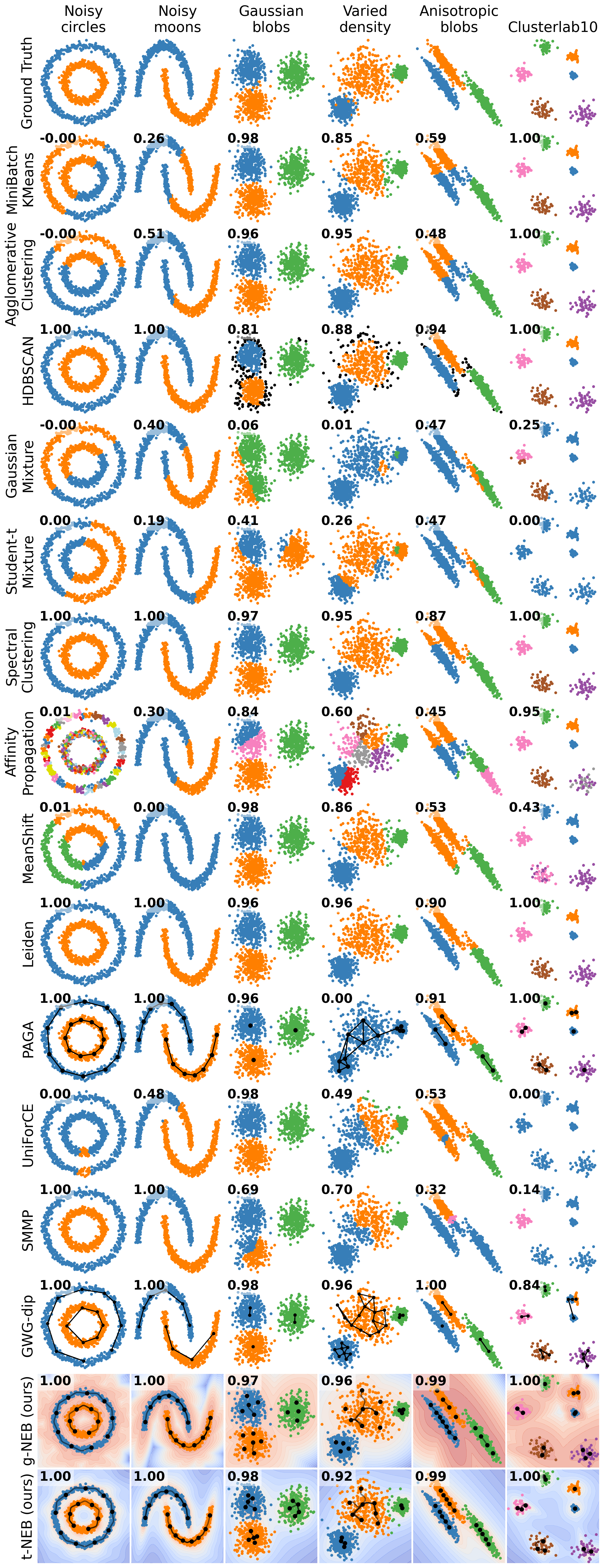}
    \caption{Clustering performance (ARI) on 2D datasets, the best of ten runs is shown. 
    Our NEB-based method correctly clusters all datasets. %
    Background in NEB columns shows the density landscape induced by the mixture model.
    Red = High Density, Blue = Low Density.}
    \label{fig:2d-datasets-app}
\end{figure*}

Multiple clustering algorithms, including g-NEB and t-NEB, show strong overall performance on the 2D toy datasets (\cref{fig:2d-datasets-app}).
Performance on the weakest dataset for t-NEB (`varied density' with ARI of 0.92) can be even further increased by allowing the underlying TMM to optimize the degrees of freedom $\nu$ instead of fixing them at 1.
Since this was detrimental on high-dimensional datasets, we sticked with the default value.
Also GWG-dip, Leiden, and spectral clustering performed well on these toy datasets.

For the other algorithms we observe the following:
First, we can clearly distinguish density-based from non-density-based algorithms by the performance on circles and moons where multiple methods fail.
Also the anisotropic blobs were able to tell apart clustering algorithms.
For example, agglomerative clustering with Ward's linkage additionally fails on the anisotropic blobs, because Ward's method minimizes the total within-cluster variance and is thus strongly biased towards round clusters. 
Also Leiden, despite being a state-of-the-art clustering algorithm, underperforms on anisotropic blobs by merging part of the green cluster into the orange one. 
This issue, common to algorithms relying on nearest-neighbor distances, arises from sensitivity to noise and low-probability connections, which density models help mitigate. 
HDBSCAN performs well on density-based datasets but requires an unusually large `min\_points' parameter to split Gaussian blobs, increasing noise labels. 
Overall, only our methods t-NEB and g-NEB, as well as Leiden, HDBSCAN, and GWG-dip perform almost perfectly on all 2D datasets.

\begin{figure*}[t]
    \centering
    \includegraphics[width=\textwidth]{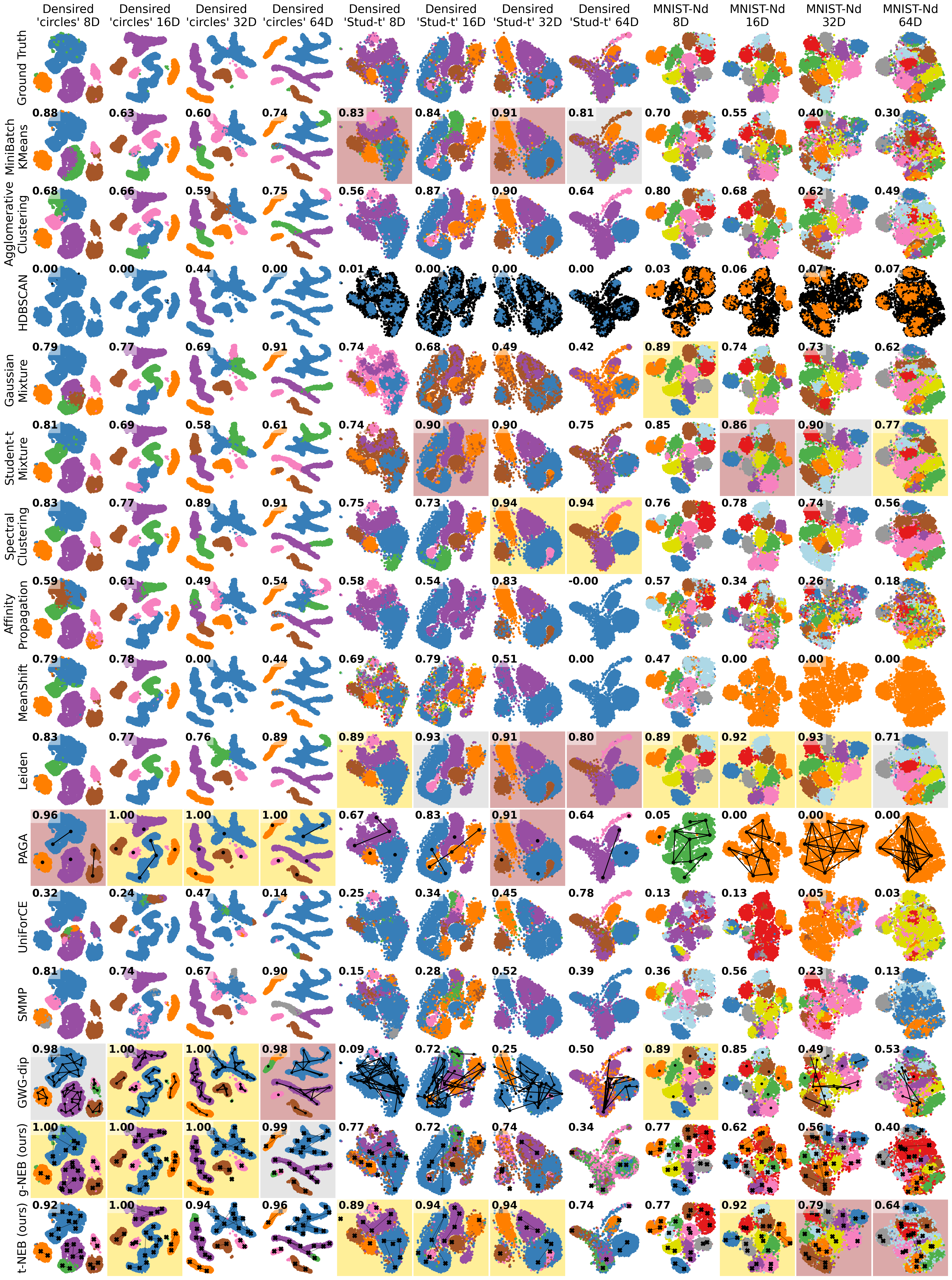}
    \caption{Clustering performance (ARI) on higher-dimensional datasets (best of ten runs).
    We use t-SNE \citep{van2008visualizing} to visualize high-dimensional datasets.
    We highlight the \colorbox{yellow!50}{1st}, \colorbox{gray!20}{2nd} and \colorbox{red!30}{3rd} best performing algorithms.}
    \label{fig:hd-datasets-app}
\end{figure*}

\paragraph{Clustering performance on higher-dimensional datasets.}
When looking at the extended performance plot in \cref{fig:hd-datasets-app}, we observe that in addition to t-NEB, g-NEB, and GWG-dip, also PAGA solves the Densired `circles' datasets (but fails on MNIST-Nd) which is why it was not included in the main paper.
In general, multiple methods (HDBSCAN, MeanShift, PAGA, UniForCE) completely fail on the MNIST-Nd datasets, while many deterioate with larger dimensions.
Also the performance on Densired `circles' is usually higher than the corresponding performance on the other two datasets, indicating that the more noise and weaker density gaps in Densired `Stud-t' and MNIST-Nd are problematic for many clustering algorithms.
Overall, the state-of-the-art (non-hierarchical) clustering algorithm Leiden performs very well, as does t-NEB. 
All other hierarchical clustering algorithms are significantly worse than t-NEB.

\FloatBarrier

\begin{table*}[ht]
\centering
\caption{Clustering results on datasets from \citet{vardakas2023UniForCE}. The best values per dataset are shown in bold. Cases marked by \textdagger indicate experiments that failed due to memory/time.
For t-NEB, $k$ was estimated by the biggest jump in thresholds, taking the median when multiple roughly equal big jumps exist.
}
\label{tab:results}
\renewcommand{\arraystretch}{1.2}
\resizebox{\textwidth}{!}{
\begin{tabular}{l@{\hskip 15pt}cc@{\hskip 15pt}cc@{\hskip 15pt}cc@{\hskip 15pt}cc@{\hskip 15pt}cc@{\hskip 15pt}cc@{\hskip 15pt}cc@{\hskip 15pt}cc}
\toprule
 & \multicolumn{2}{c@{\hskip 15pt}}{\textbf{Optdigits}} & \multicolumn{2}{c@{\hskip 15pt}}{\textbf{Pendigits}} & \multicolumn{2}{c@{\hskip 15pt}}{\textbf{EMNIST-BD}}  & \multicolumn{2}{c@{\hskip 15pt}}{\textbf{EMNIST-BL}} & \multicolumn{2}{c@{\hskip 15pt}}{\textbf{EMNIST-M}}
 & \multicolumn{2}{c@{\hskip 15pt}}{\textbf{Waveform-v1}} & \multicolumn{2}{c@{\hskip 15pt}}{\textbf{Mice Protein}} \\
 & \( k \) & AMI & \( k \) & AMI & \( k \) & AMI & \( k \) & AMI
 & \( k \) & AMI & \( k \) & AMI & \( k \) & AMI \\
\midrule
RCC         & 19$\pm$0  & \textbf{0.87} & 46$\pm$0 & 0.75 & 82$\pm$0 & 0.74 & 82$\pm$0 & 0.52 & \textdagger & \textdagger & \textbf{3$\pm$0}  & \textbf{1.00} & 54$\pm$0 & 0.52  \\
UniForCE    & \textbf{11$\pm$1}  & 0.85 & 17$\pm$1 & 0.78 & \textbf{10$\pm$1} & 0.86 & \textbf{12$\pm$1} & 0.74 & 13$\pm$1 & 0.84 & \textbf{3$\pm$0}  & \textbf{1.00} & \textbf{8$\pm$0} & \textbf{0.93}  \\
t-NEB       & \textbf{11$\pm$1}  & 0.76 & \textbf{17$\pm$0} & \textbf{0.80} & \textbf{10$\pm$1} & \textbf{0.92} & \textbf{12$\pm$1} & \textbf{0.85} & \textbf{10$\pm$2} & \textbf{0.94} & \textbf{3$\pm$1}  & \textbf{1.00}  & 13$\pm$12  & 0.53 \\
\midrule
Ground truth & 10 & - & 10 & - & 10 & - & 10 & - & 10 & - & 3 & - & 8 & -  \\
\bottomrule
\end{tabular}
}

\end{table*} 
\paragraph{Evaluating on datasets from \citet{vardakas2023UniForCE}.}
\label{app:uniforcedata}
In addition to the comparison on our datasets, we use the datasets from Table 2 in the UniForCE paper \citep{vardakas2023UniForCE} to evaluate against both UniForCE and RCC \citep{shah2017robust}.
For Dataset statistics, see \cref{tab:datasets} `UniForCE'.
Our method outperforms both RCC and UniForCE on most datasets, but underperforms on the `Mice Protein Expression' and `Optodigits' dataset due to too few samples for its dimensionality, limiting our method's ability to effectively estimate the probability density of the data (\cref{tab:results}). 
We did not extensively tune our method, using 25 initial clusters and default regularization strength, adjusting only when stronger regularization was needed to prevent filtering out too many components.

\paragraph{Evaulating the underlying mixture model: GMM vs TMM.}
\label{app:extendedResultsTMMvsGMM}

\begin{figure}[H]
    \centering
    \includegraphics[width=0.7\linewidth]{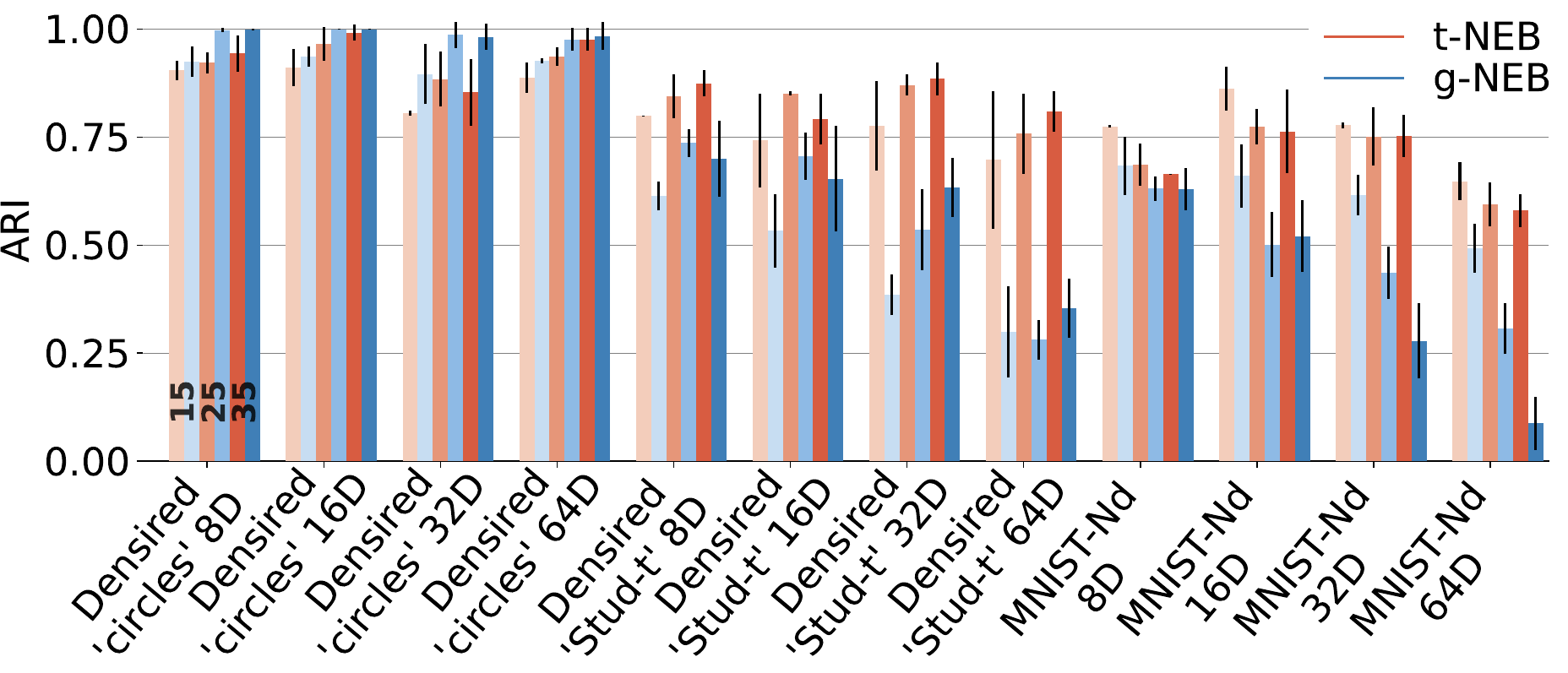}
    \caption{
    t-NEB (red) vs g-NEB (blue) ARI with 15, 25 and 35 initial mixture components (different shades). 
    Especially on the more noisy and real-world-like datasets Densired `Stud-t' and MNIST-Nd, TMM-based models work better.}
    \label{fig:tmm-gmm}
\end{figure}

Mixture modeling typically uses Gaussian distributions unless assumptions about the data suggest otherwise. 
However, real-world data is often not normally distributed and includes outliers. 
To address this, we use Student's $t$-distributions, a generalization of the Gaussian distribution. 
As the degrees of freedom increase, the $t$-distribution converges to a Gaussian while lower degrees of freedom allow for heavier tails.
In our experiments, Student's $t$ mixture models tended to perform better which we attribute to the fact that the distribution is less susceptible to outliers.
This can be seen in \cref{fig:tmm-gmm} where on the Densired `Stud t' and MNIST-Nd datasets the red bars are visibly higher than the blue ones, independent of the chosen dimension.
We also see that on MNIST-Nd it is detrimental to fit more clusters during the overfitting step (especially for the GMM models) while on Densired `Stud t' the opposite is the case - more initial components tends to lead to better overall performance.

\section{Hierarchy of real-world transcriptomic cell-types}
\label{app:realhierarchy}

\begin{figure}
    \centering
    \includegraphics[width=\linewidth]{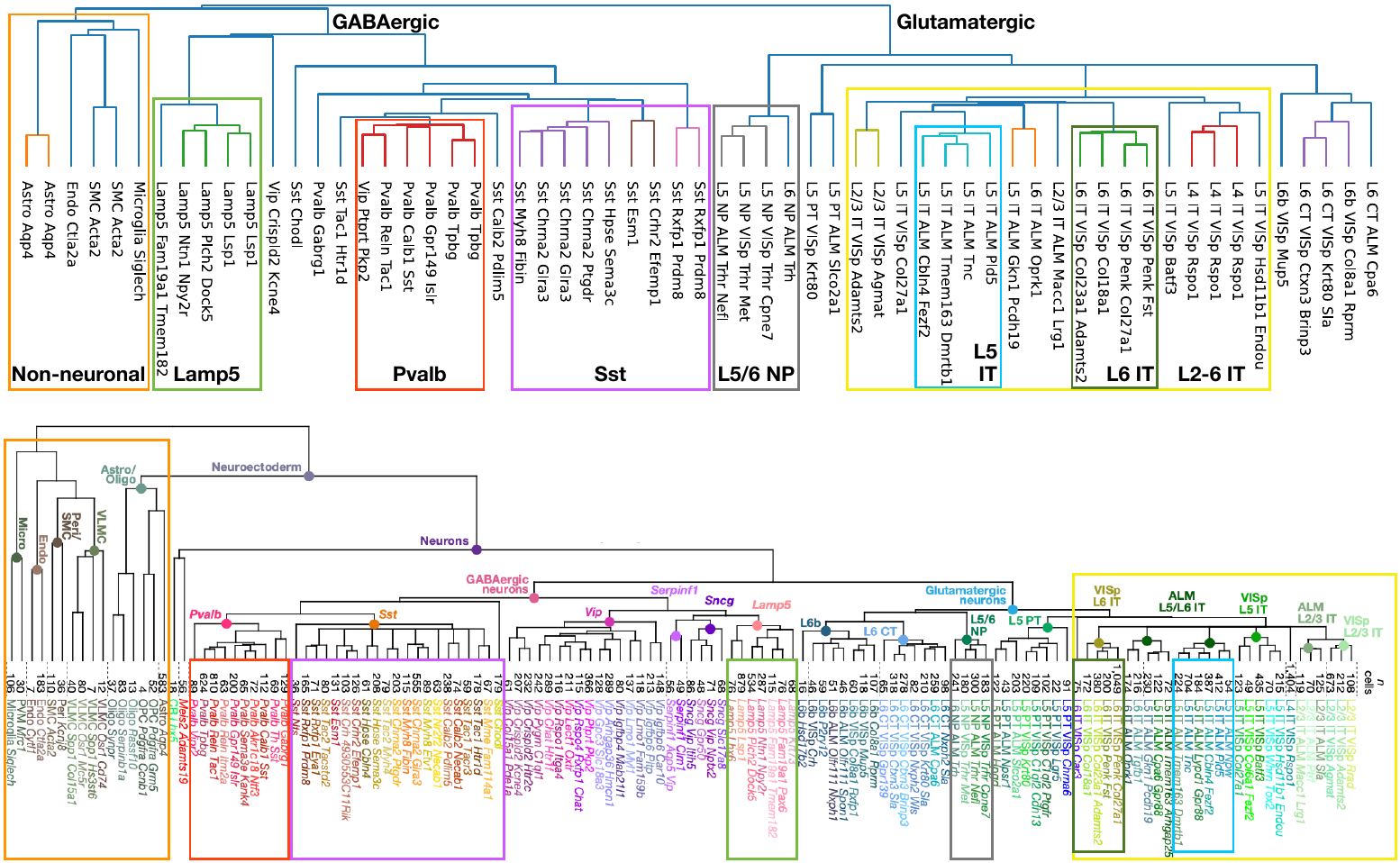}
    \caption{Top: Hierarchy produced by t-NEB (labels matched after clustering). 
    Bottom: Original  hierarchy from \citet{tasic2018shared}, figure adjusted from Fig. 1 in their paper.
    }
    \label{fig:tasic-app}
\end{figure}

We use a widely accepted dataset from \citet{tasic2018shared} and preprocess it following \citet{kobak2019art}, with 23,822 cells across 133 cell types.
To reduce noise, we exclude cell types represented fewer than 100 times, resulting in 20,502 cells and 75 cell types.
To mimic an exploratory setting, we then apply t-NEB to the 50-dimensional data without hyperparameter tuning (75 initial clusters, 0.01 regularization), obtaining 61 components.
Predicted clusters are matched to ground truth labels by assigning each cluster the most frequent label within it.
If a label was already assigned to another cluster, we select the next most frequent unused label.
Due to label imbalance, some predicted clusters share the same label, yielding 53 uniquely labeled clusters.

Most of the \emph{L6b} cells were filtered out as they had less than 100 cells, which results in t-NEB merging the \emph{L6b} and \emph{L6CT} groups (very right of the top plot in \cref{fig:tasic-app}).
For \emph{L5 PT}, t-NEB identifies only two groups instead of five non-filtered ones, but these 2 are correctly placed in the hierarchy on several levels.
Next, to the left we have 3 red groups with identical \emph{`L4 IT VISp Rspo1'} label. 
This happens as \emph{`L4 IT VISp Rspo1'} is one of the most dominant cell types across the whole dataset, having 1404 cells (column 7 from the right, bottom subplot in \cref{fig:tasic-app}) and it forms its own subcluster in the original hierarchy as well, staying closest to \emph{`L4 IT VISp Batf3'} and \emph{`L4 IT VISp Hsd11b1 Endou'}, which t-NEB merges it with. 
The same holds for \emph{`Astro Aqp4'} in non-neuronal cells (orange) which got assigned two groups as it is a dominant cell type ($n=583$).
We do not get any \emph{`Sncg'} subgroups, as all cells in \emph{`Sncg'} are present less than 100 times.
Surprisingly, we get most of the \emph{`Vip'} cells matched to the \emph{`Vip Ptprt Pkp2'} group, resulting in $n=12,722$ cells in this group.
This likely happens as the number of initial clusters is relatively low.
Additionally, such imbalanced grouping likely causes a huge component, which results in a low density gap between the \emph{`Vip Ptprt Pkp2'} group and lead to it being is incorrectly merged with the \emph{`Sst'} group (red box).
Overall, despite some differences from the original paper, t-NEB effectively captures the hierarchical organization of transcriptomic cell types, making it a valuable tool for exploratory analysis of high-dimensional biological data.

\FloatBarrier

\section{Hyperparameters}
\label{app:hyperparameters}

\begin{table}[h]
\begin{center}
\caption{t/g-NEB Hyperparameters. Key hyperparameters in \textbf{bold}.}
\label{tab:hyperparameters}
\begin{tabular}{rlr}
\toprule
& \textbf{Parameter} & \textbf{Default} \\
\midrule
\multirow{2}{*}{\rotatebox[origin=c]{90}{new}} & \textbf{Number of clusters (overclustering)} & 25 \\
& Number of kNN Neighbors for which NEB paths are computed & 10 \\
\midrule
\multirow{7}{*}{\rotatebox[origin=c]{90}{{Mixture Model}}}& \textbf{TMM regularization} & 1e-4 \\
& TMM degrees of freedom (df) & 1 (fixed) \\
& Covariance type & `full' \\
& EM steps & 1000 \\
& Initialization & `kmeans' \\
& Number of initializations (n\_init) & 20\\
& Minimum cluster size & 10 \\
& Maximum cluster elongation & $500d$ \\
\midrule
\multirow{3}{*}{\rotatebox[origin=c]{90}{NEB}}& 
Number of steps for NEB & 200\\
& Number of points on the NEB path & 100 \\
& Number of points to evaluate final NEB path & 1024 \\
\bottomrule
\end{tabular}
\end{center}
\end{table} 
We list the hyperparameters of our algorithm in \cref{tab:hyperparameters}.
There are three groups of hyperparameters, belonging to t-NEB/g-NEB, mixture model fitting, and NEB computation.
As stated in the main paper, the two important hyperparameters are the number of clusters for overclustering and in the case of TMM the regularization parameter.
Mixture model fitting has additional parameters such as the number of EM steps, initialization, tolerance, and covariance type, all of which impact the base clustering on which our merging procedure happens.

\subsection{t-NEB specific hyperparameters}

\paragraph{Number of initial clusters during overclustering.}

\begin{figure}[t]
    \centering
    \includegraphics[width=0.9\linewidth]{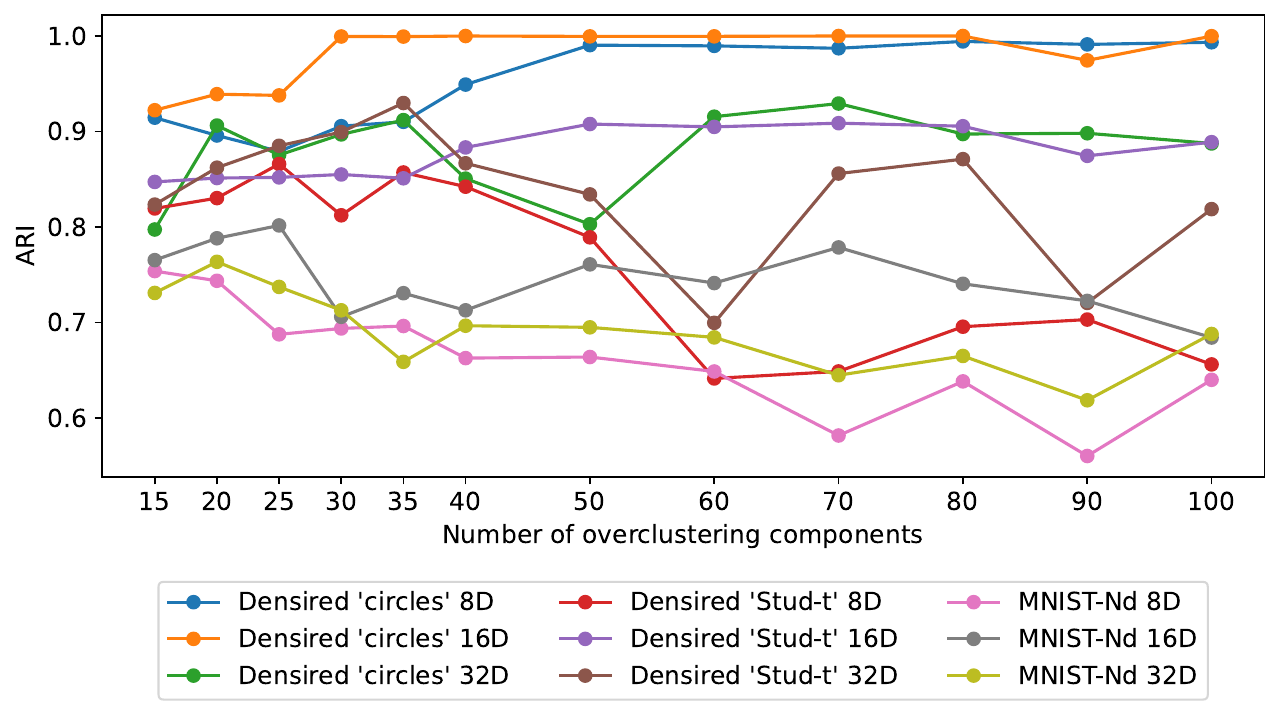}
    \caption{Varying the number of initial components during overclustering, showing the mean ARI of four t-NEB models. While some datasets profit from large numbers of initial components, a moderate number of 20-25 is a reasonable choice.}
    \label{fig:num_components}
\end{figure}

\begin{figure}[t]
    \centering
    \includegraphics[width=0.7\linewidth]{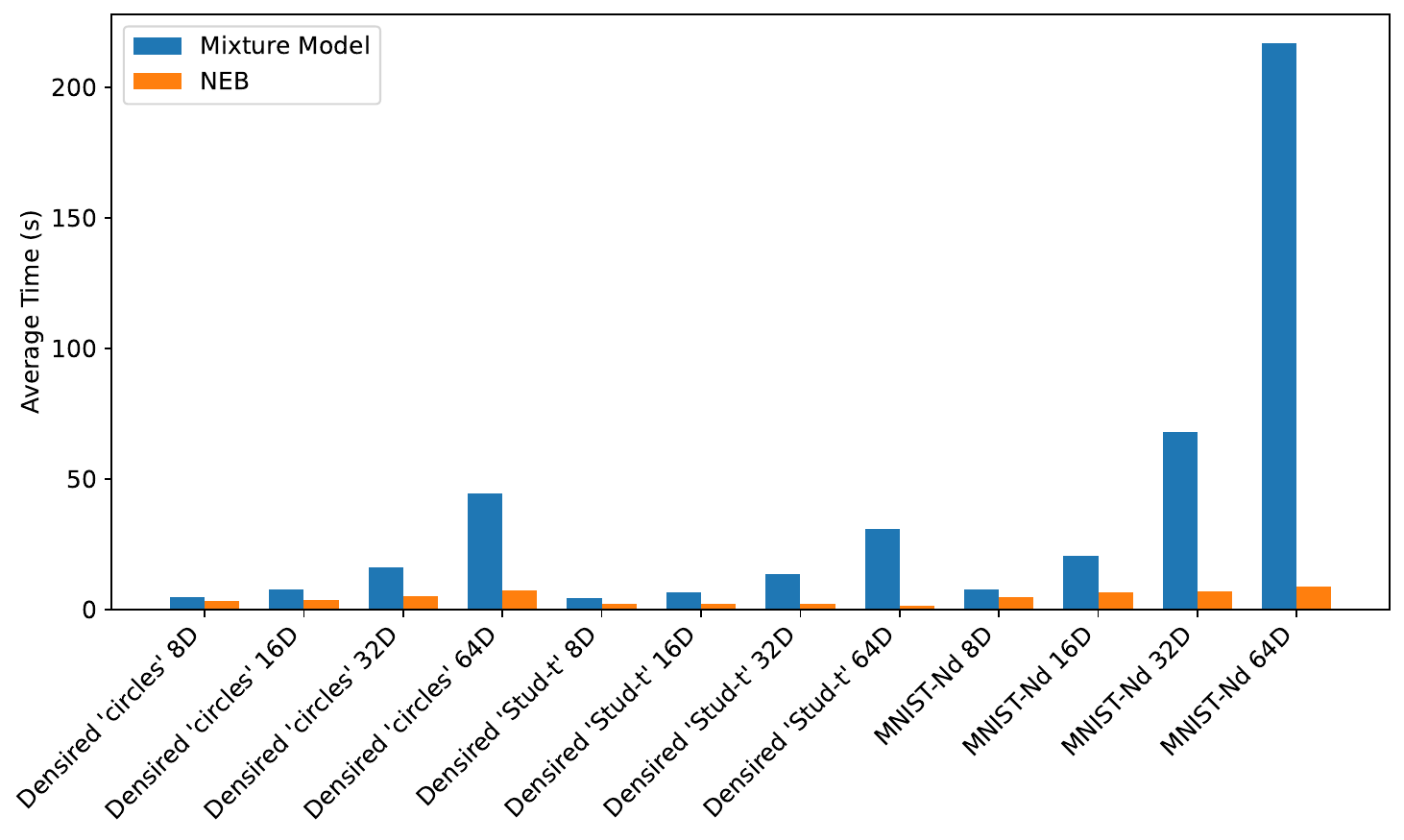}
    \caption{Runtime of t-NEB analyzed. The mixture model fitting dominates the total runtime.}
    \label{fig:runtime}
\end{figure}

The number of clusters during overclustering is the main hyperparameter of our method.
Choosing the right value for this parameter often makes the difference between a reasonable and a very good clustering.
In \cref{fig:num_components} we plot average ARI of four models against the number of overclustering components.
In the experiment we skip the 64D variants of each dataset due to runtime constraints (see \cref{fig:runtime}).

We observe that there is no clear trend. While some datasets (especially Densired `Circles') profit from large numbers of initial components, a smaller number of initial clusters in the range of 20 to 25 is a reasonable choice.
In our experiments, we use 20 initial clusters for MNIST-Nd and 25 for the Densired datasets.
Further increasing this number for Densired might have improved clustering performance while the choice was close to optimal for MNIST-Nd.

\begin{figure}
    \centering
    \includegraphics[width=0.9\linewidth]{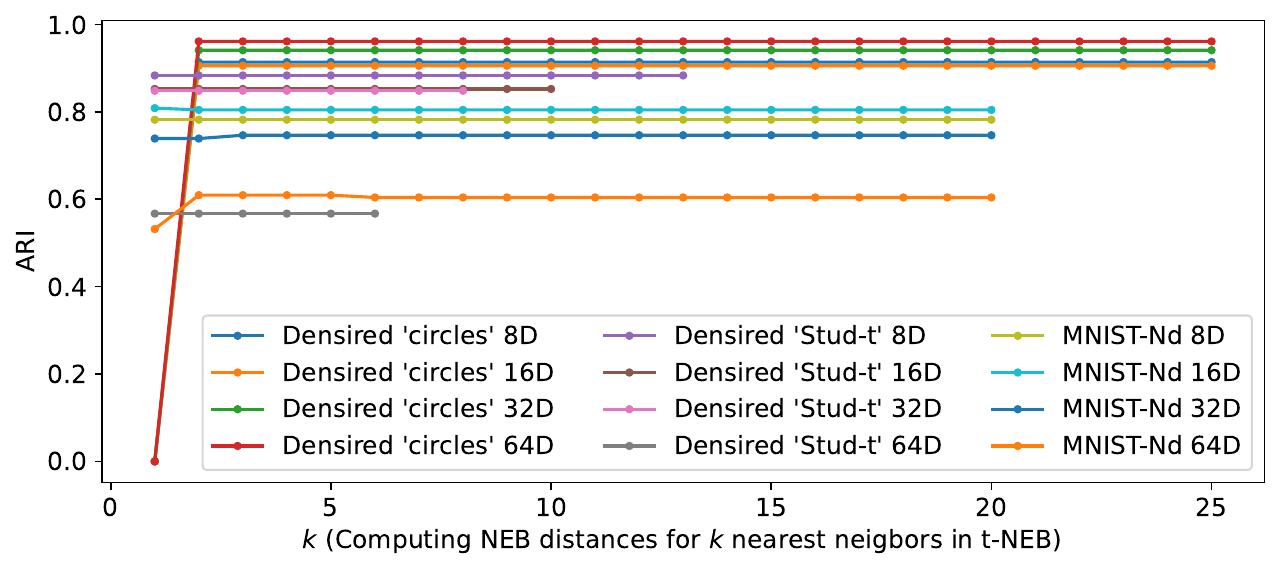}
    \caption{Varying the number of $k$ neighbors for which NEB paths are computed.
    Maximum number of neighbors is bounded by the number of mixture components (Densired: \(m=25\), MNIST-Nd: \(m=20\); filtering of small and elongated components took place for Densired `Stud-t', see \cref{fig:filtering_scatterplot}).
    Performance saturates for $k=3$. 
    }
    \label{fig:ablation_k_neighbors}
\end{figure}

\paragraph{K-neighbors.} 
Long high-density paths typically pass through other cluster centers such that computing NEB paths for close (w.r.t. Euclidean distance) clusters followed by the MST computation gives identical results and is faster\footnote{In theory, it might even improve results if NEB is getting stuck in local optima. We did observe such cases for low numbers of clusters on the moons dataset, even though they did not influence the merging order.}.
For example, on Densired ‘circles’ 32D (30 TMM components), $k=1$ fails (the graph has more than six connected components), $k=2$ gives ARI 0.76, and $k\geq 3$ reaches 0.90.
We plot ARI against $k$ 
in \cref{fig:ablation_k_neighbors}.
Note that for Densired `Stud-t' components with single elements and high elongation were filtered out which results in differing maximum number of $k$ (more information on this filtering step in \cref{par:filtering}). 
The plot shows that on all datasets choosing $k=10$ is always more than enough and in most cases $k=3$ would have been enough.

\subsection{Mixture model specific hyperparameters}

\paragraph{TMM regularization.}

\begin{figure}
    \centering
    \includegraphics[width=0.9\linewidth]{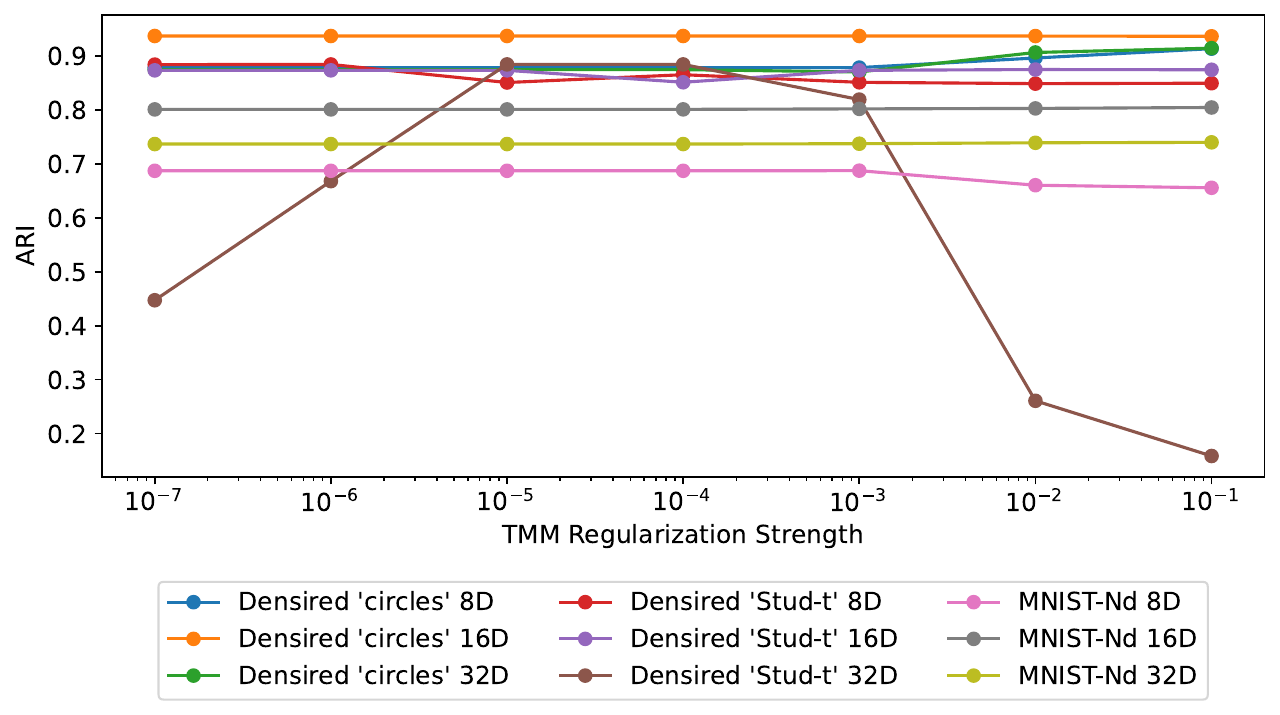}
    \caption{Effect of regularization on performance. In most cases, regularization strength is not extremely important.}
    \label{fig:regularization}
\end{figure}

The regularization parameter influences the flexibility of TMM to match the given density.
Higher regularization leads to faster convergence, especially in high-dimensional datasets, while reducing the flexibility.
On our datasets, the exact choice of the regularization parameter was barely relevant (except for one dataset Densired `Stud-t' 32D), while on the datasets used in the UniForCE paper \citep{vardakas2023UniForCE}, it made a difference.
To check the influence of the regularization parameter, we computed TMM models for different regularization strength over all datasets with up to 32 dimensions (TMM fitting of 64D is quite slow, especially for low regularization).
\cref{fig:regularization} shows the resulting ARI of t-NEB (mean among four models) against the regularization strength. 
We observe that regularization only has a low effect on the overall performance and values between $10^{-5}$ and $10^{-3}$ work for all datasets.
Our default of $10^{-4}$ is thus a reasonable choice.

\begin{figure}
    \centering
    \includegraphics[width=.5\linewidth]{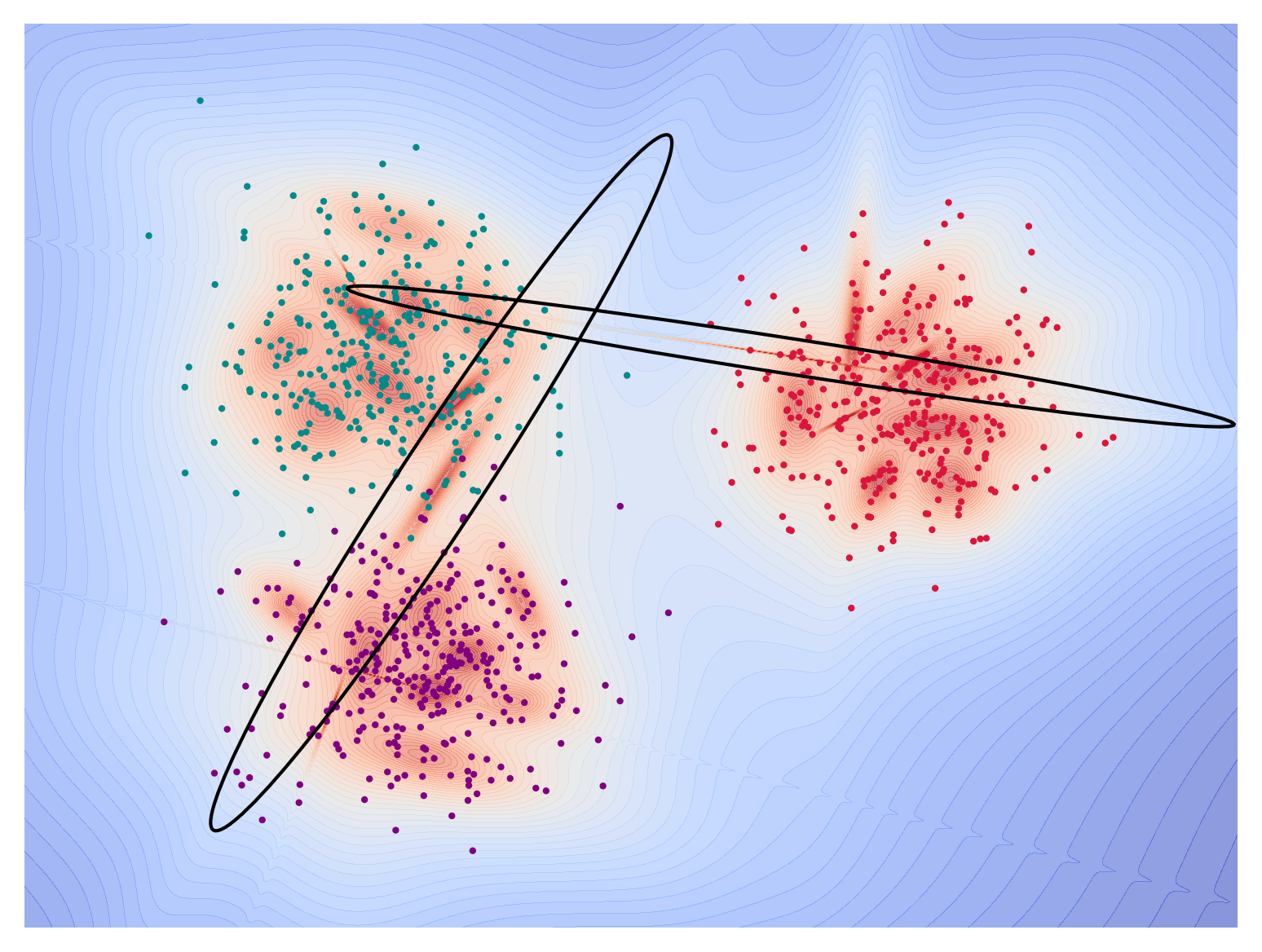}%
    \includegraphics[width=.5\linewidth]{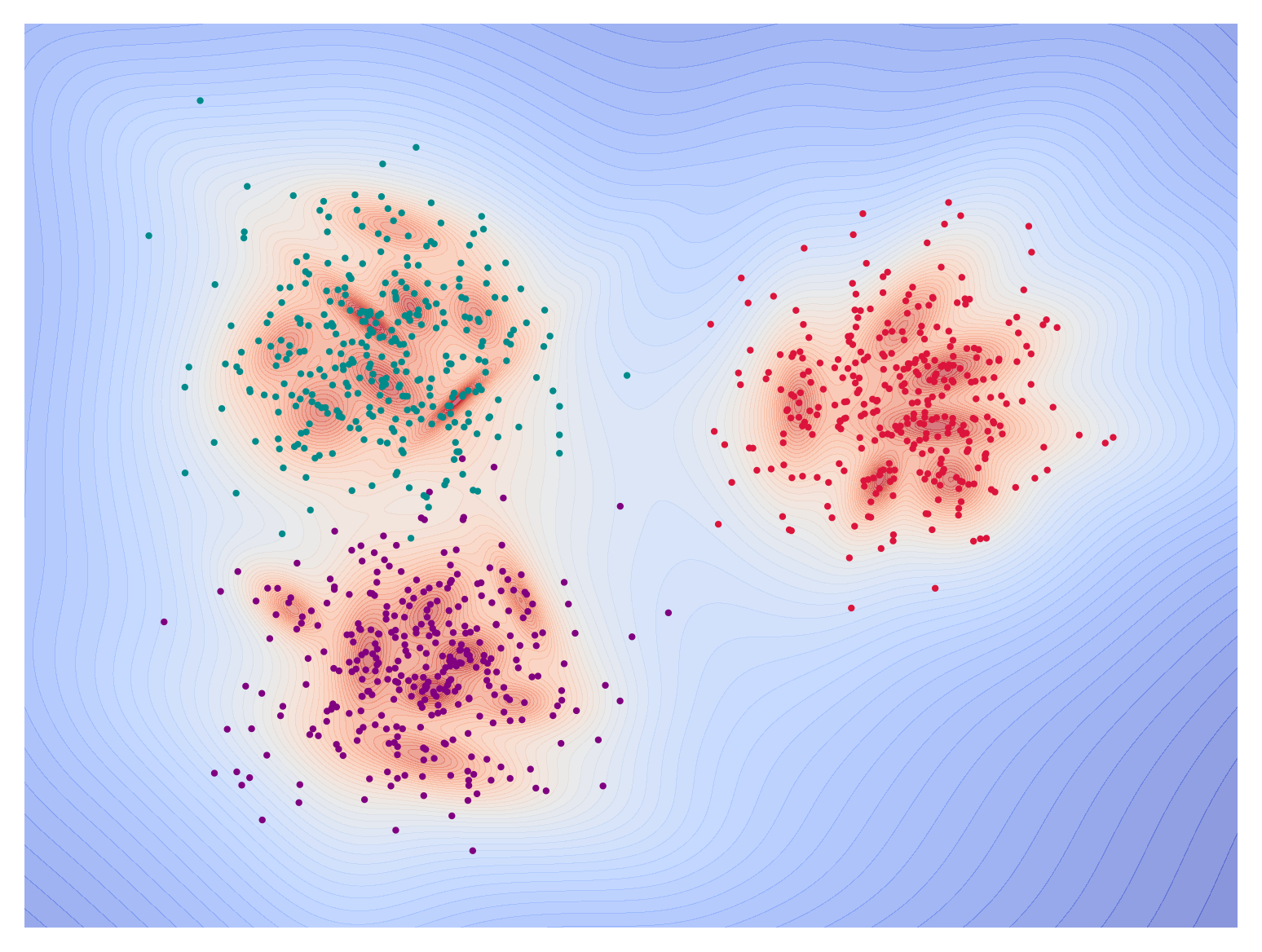}    
    \caption{Elongated clusters in unfiltered TMM serve as ``bridges'' in the density. \emph{Left:} Full density landscape of a TMM model with 30 components on 2D Gaussian blobs. Ellipses: problematic components acting as bridges. 
    \emph{Right:} Filtered density.}
    \label{fig:filtering}
\end{figure}

\begin{figure}
    \centering
    \begin{minipage}{0.7\textwidth}
        \includegraphics[width=\linewidth]{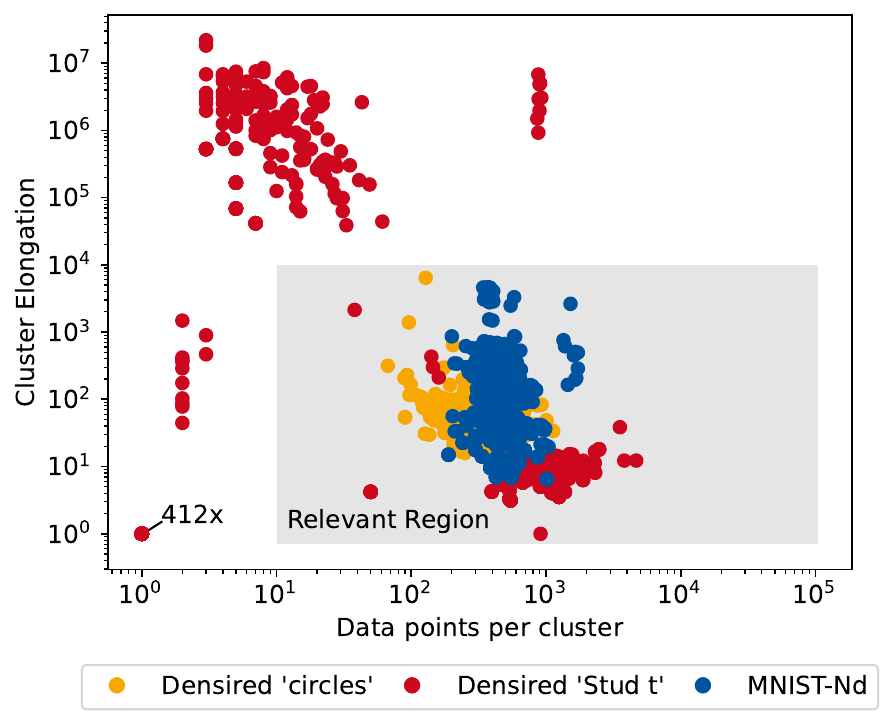}
    \end{minipage}%
    \begin{minipage}{0.3\textwidth}
    \resizebox{!}{4cm}{
\begin{tabular}{cc}
\toprule
\textbf{Cluster size} & \textbf{Elongation} \\
\midrule
47    & 102228    \\
1     & 1         \\
\cellcolor{black!10} {1876}  & \cellcolor{black!10} {6}         \\
1     & 1         \\
1     & 1         \\
\cellcolor{black!10} {1219}  & \cellcolor{black!10} {7}         \\
25    & 277239    \\
1     & 1         \\
30    & 140863    \\
\cellcolor{black!10} {552}   & \cellcolor{black!10} {6}         \\
1     & 1         \\
1     & 1         \\
\cellcolor{black!10} {2318}  & \cellcolor{black!10} {8}         \\
\cellcolor{black!10} {1829}  & \cellcolor{black!10} {8}         \\
1     & 1         \\
1     & 1         \\
2     & 752       \\
1     & 1         \\
\cellcolor{black!10} {910}   & \cellcolor{black!10} {9}         \\
8     & 2879403   \\
1     & 1         \\
1     & 1         \\
1     & 1         \\
\cellcolor{black!10} {1154}  & \cellcolor{black!10} {7}         \\
18    & 536639    \\
\bottomrule
\end{tabular}
}
    \end{minipage}%
    
    \caption{\textbf{Plot}: For each cluster we plot the number of datapoints against the elongation for the three main datasets. We filter out clusters with only one element (singletons) and clusters with a high elongation (outside of the \colorbox{black!10}{relevant region}). 
    Only for Densired `Stud $t$' filtering took place with an emphasis on singletons: 412 out of 1000 base clusters where singletons and thus filtered. Another 167 were filtered due to their elongation. For the other datasets, no filtering was needed. \textbf{Table}: Cluster sizes and elongations before filtering on Densired `Stud t' 32D, clusters kept during filtering are marked in \colorbox{black!10}{gray}.}
    \label{fig:filtering_scatterplot}
\end{figure}

\paragraph{Filtering: Minimum cluster size and maximum cluster elongation.}
\label{par:filtering}
When fitting Gaussian or Student's $t$ mixture models with the EM algorithm, issues with overly-elongated components often arise (see \cref{fig:filtering}).
This happens especially in high dimensions where components may have fewer assigned points than degrees of freedom. 
To avoid these issues we use ridge regularization with strength $10^{-4}$ (see the previous paragraph for the strength of regularization).
Even with regularization, some clusters remain overly elongated and on Densired `Stud t' a large number of singletons are fit.
Out of 1000 base clusters among TMM models fit on Densired `Stud t' (using 25 clusters per model), 412 are singletons and another 167 are highly elongated, summing up to over half of the clusters.
In \cref{fig:filtering_scatterplot} we see that in many cases, highly-elongated clusters are also small.
We thus discard clusters with less than $M=10$ points (similar to HDBSCAN) and components with a maximal elongation greater than $E=500 \cdot d$, where elongation is defined as the ratio of the largest and smallest eigenvalues of the scale matrix.
Since the number of points affected by these changes is small, we argue that this filtered clustering approximates the original one while being well-behaved.

We note that elongated components are potentially problematic for t-NEB because they can act as `bridges' between otherwise not connected clusters and the NEB algorithm finds those also in practice. 
An example for such bridges is given in \cref{fig:filtering}.

Filtering is the main reason why mixture model fitting sometimes results in less mixture components than expected (and initially started with), e.g. in \cref{fig:realhierarchy} and also in \cref{fig:ablation_k_neighbors}.
Components after filtering are highlighted as \colorbox{black!10}{relevant region}) in the plot and in the table in (\cref{fig:filtering_scatterplot}).
In our main experiments, filtering took place only for Densired `Stud t' as described above.

\subsection{NEB specific hyperparameters}
\paragraph{Number of steps for NEB.} 
\begin{figure}
    \centering
    \begin{minipage}{0.5\textwidth}
        \includegraphics[width=\textwidth]{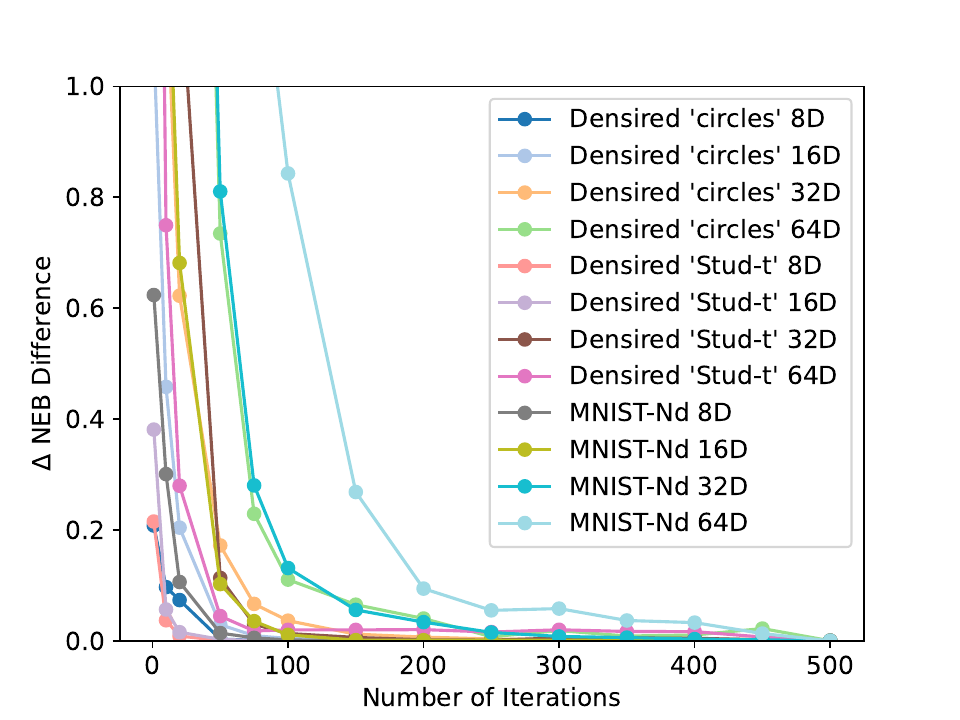}
    \end{minipage}%
    \begin{minipage}{0.5\textwidth}
        \includegraphics[width=\linewidth]{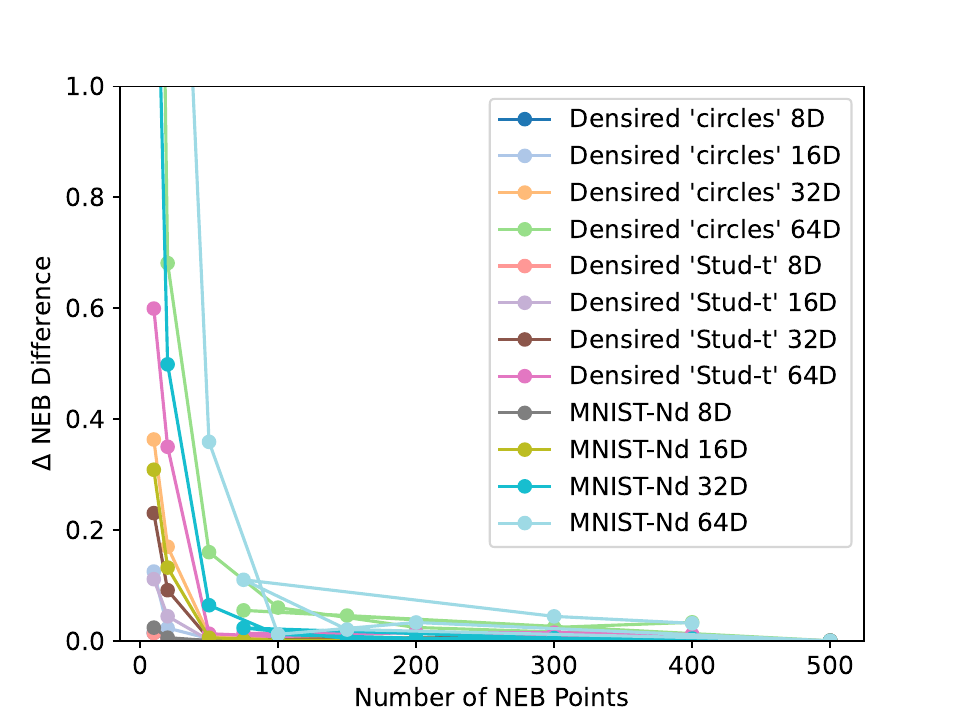}
    \end{minipage}
    \caption{\textbf{Left:} Varying number of NEB optimization steps. After 200 iterations the distance matrix has converged on all datasets.
    \textbf{Right:} Varying NEB path length. Optimizing paths of length 100 is sufficient for all datasets. 
    This experiment used 500 NEB iterations to ensure convergence.
    }
    \label{fig:NEB-iterations}
    \label{fig:NEB-pathlength}
\end{figure}
In \cref{fig:NEB-iterations} (left) we show the effect of running NEB for longer.
We can observe that on all datasets the computed distances converge after at most 200 iterations.
We thus expect 200 iterations to generalize well, especially since minor changes in the distance matrix typically do not impose a difference in the merging order (not shown here) and thus do not impact the computed clustering.

\paragraph{Number of points on the NEB path.}
The second core hyperparameter of NEB is the number of points on each path that are optimized to follow the maximum energy path.
In \cref{fig:NEB-pathlength} (right) we show the effect of optimizing different numbers of points along each path. 
Here, we use 500 NEB iterations to ensure convergence and compare the computed adjacency matrices to the one for the maximum number of points (500 points per path). 
We plot the Frobenius Norm of the difference.
We observe that at around 100 points the results converge.

\begin{figure}[H]
    \centering
    \includegraphics[width=0.55\linewidth]{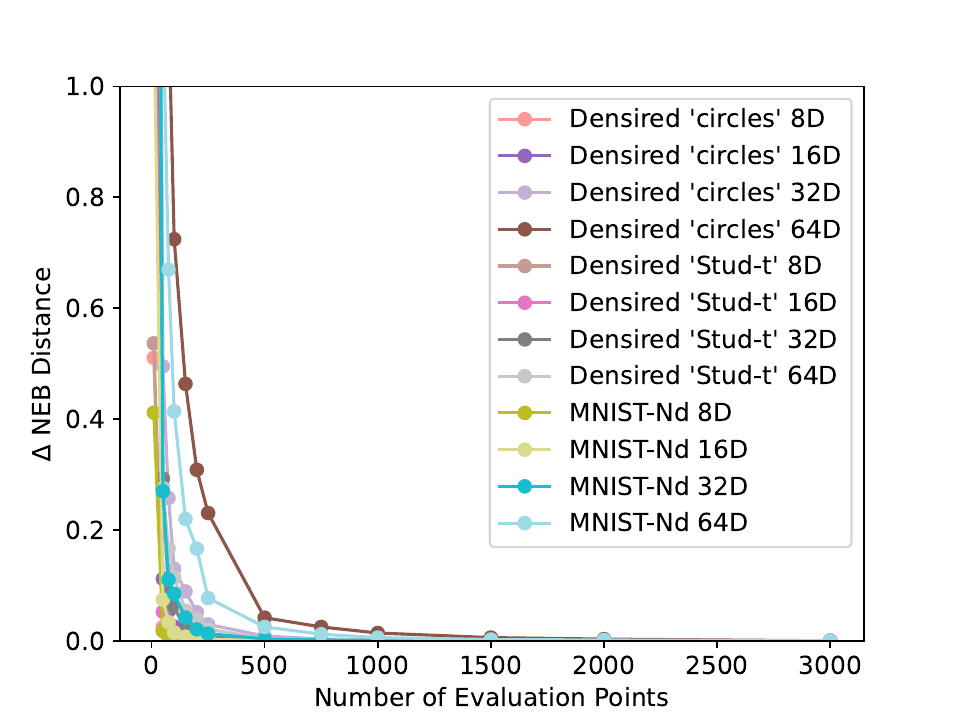}
    \caption{Average difference in computed NEB distance based on the number of evaluation points. Low sampling rates lead to large errors in the computed distances while a sampling rate above 500 is sufficient to consistently estimate the minimum density between cluster centers. We choose a sampling rate of 1000.}
    \label{fig:ablationNebEvalPoints}
\end{figure}

\paragraph{NEB Evaluation Points.}
When computing the distance between two cluster centers, we compute the probability of the TMM model for the NEB path between them.
Especially for small path lengths and large distances between cluster centers, this may lead to inaccurate estimation of the minimum distance along the path since our samples might be too sparse and simply miss the low areas.
This can be avoided by increasing the sampling rate when evaluating the distance between cluster centers.
In our experiments we sampled 1024 points between any two cluster centers.
In order to verify that this value is indeed sufficient, we computed the distances for all paths of each dataset with different number of NEB evaluation points (i.e. sampling rates). 
\cref{fig:ablationNebEvalPoints} shows the difference in the computed distance against the final distance (on average over all paths) for each dataset.
We observe that with less than 200 samples, the computed distances might differ significantly, while sampling 500 or 1000 points leads to consistent performance over all datasets.
This indicates that sampling 1024 points as we did in our experiments is indeed a valid choice.

\FloatBarrier

\section{Merging Strategies}

\begin{table*}[!ht]
    \centering
    \resizebox{\textwidth}{!}{
    \begin{tabular}{ll|ll|lll|lll}
    \toprule
        Dataset & Dim & Oracle & NEB & Euclidean & $k$-means & Dip & Euclidian & $k$-means & Dip \\ 
         &  &  &  & with & with & with &  w/o &  w/o &  w/o \\ \midrule

        \multirow{4}{1.3cm}{Densired `Circles'} & 8 & 0.99 ± 0.00 & 0.92 ± 0.02 & 0.83 ± 0.05 & 0.87 ± 0.06 & 0.17 ± 0.05 & 0.21 ± 0.00 & 0.22 ± 0.01 & 0.19 ± 0.01 \\ 
       ~&16 & 1.00 ± 0.00 & 0.96 ± 0.04 & 0.65 ± 0.02 & 0.67 ± 0.05 & 0.12 ± 0.08 & 0.22 ± 0.01 & 0.21 ± 0.01 & 0.20 ± 0.01 \\ 
        ~&32 & 1.00 ± 0.00 & 0.89 ± 0.06 & 0.70 ± 0.11 & 0.74 ± 0.11 & 0.12 ± 0.06 & 0.22 ± 0.01 & 0.22 ± 0.01 & 0.20 ± 0.00 \\ 
        ~&64 & 1.00 ± 0.00 & 0.94 ± 0.02 & 0.79 ± 0.06 & 0.74 ± 0.11 & 0.15 ± 0.05 & 0.22 ± 0.01 & 0.23 ± 0.01 & 0.20 ± 0.01 \\ \hline
        \multirow{4}{1.1cm}{Densired `Stud-t'} & 8 & 0.88 ± 0.00 & 0.85 ± 0.05 & 0.86 ± 0.04 & 0.00 ± 0.00 & 0.21 ± 0.02 & 0.43 ± 0.03 & 0.43 ± 0.06 & 0.35 ± 0.01 \\ 
        & 16 & 0.94 ± 0.00 & 0.85 ± 0.00 & 0.73 ± 0.09 & 0.00 ± 0.00 & 0.34 ± 0.09 & 0.50 ± 0.05 & 0.48 ± 0.04 & 0.39 ± 0.03 \\ 
        ~&32 & 0.94 ± 0.00 & 0.87 ± 0.02 & 0.87 ± 0.02 & -0.00 ± 0.00 & 0.42 ± 0.09 & 0.65 ± 0.11 & 0.63 ± 0.15 & 0.46 ± 0.06 \\ 
        ~&64 & 0.78 ± 0.03 & 0.66 ± 0.09 & 0.66 ± 0.09 & -0.00 ± 0.00 & 0.54 ± 0.14 & 0.63 ± 0.07 & 0.81 ± 0.07 & 0.55 ± 0.12 \\ \hline
        \multirow{4}{1.1cm}{MNIST-Nd} & 8 & 0.89 ± 0.01 & 0.68 ± 0.05 & 0.77 ± 0.03 & 0.64 ± 0.07 & 0.31 ± 0.07 & 0.55 ± 0.01 & 0.49 ± 0.01 & 0.50 ± 0.01 \\ 
        ~&16 & 0.92 ± 0.01 & 0.78 ± 0.04 & 0.33 ± 0.13 & 0.35 ± 0.12 & 0.32 ± 0.06 & 0.57 ± 0.01 & 0.45 ± 0.01 & 0.53 ± 0.00 \\ 
        ~&32 & 0.89 ± 0.02 & 0.76 ± 0.06 & 0.17 ± 0.03 & 0.13 ± 0.06 & 0.26 ± 0.06 & 0.53 ± 0.02 & 0.38 ± 0.01 & 0.50 ± 0.01 \\ 
        ~&64 & 0.65 ± 0.04 & 0.55 ± 0.06 & 0.10 ± 0.02 & 0.08 ± 0.02 & 0.14 ± 0.06 & 0.39 ± 0.02 & 0.32 ± 0.01 & 0.38 ± 0.02 \\ \bottomrule
    \end{tabular}
    }
    \caption{Merging strategies, extended version of table in \cref{fig:join_strategies}. Strategies `with' recompute cluster centers after each merge (relevant for e.g. Euclidean distance which is computed between centers).
    Strategies `w/o' compute all distances at once and perform the $c$ best merges from this list (i.e. one-shot merging). On the (ball-shaped) Gaussian datasets, simple Euclidean merging with recomputation performs best, for the other datasets NEB dominates the other merging strategies.
    Since $k$-means uses different underlying clusters, it can be better than the oracle.}
    \label{tab:merging_strategies_appendix}
\end{table*}

In addition to the merging strategies in the main paper, we checked additional strategies. %
The columns Oracle, NEB, Euclidean `with', and Dip `with' are also presented in the main paper.
The `with' variants of the algorithms recompute the cluster centers after each merge (the points are still just merged, i.e. no new fitting of mixture models takes place).
In contrast, the methods `w/o' do not recompute the distances between each merge, meaning that they compute all pairwise distances, sort them, and then perform as many merges from this list until the target number of clusters is reached.

The additional column $k$-means uses a $k$-means overclustering instead of the mixture model overclustering used by the other columns.
It then merges clusters based on Euclidean distance.

We see that NEB clearly outperforms all other merging strategies, especially in higher dimensions.
However, there are some cases of non-Gaussian datasets where Euclidean merging is on par or even better than  NEB, e.g. Densired `Stud-t' in 32D and 64D as well as MNIST-Nd 8D.
While $k$-means works relatively well on Densired `circles' and MNIST-Nd 8D, it outright fails on Densired `Stud-t' across all tested dimensions.
$k$-means merging on Densired `Stud-t' could be improved by removing centers merging, in contrast to Euclidean distance, which prefers merging the centers.
Removing merging centers also noticably improves dip-statistics merging but it never reaches close to NEB performance.
Overall we see that 
NEB outperforms all other approaches.

\FloatBarrier
\section{Stability Evaluation}
\label{app:stability}

\begin{figure}
    \centering
    \begin{subfigure}[b]{0.23\linewidth}
        \includegraphics[height=3.6cm]{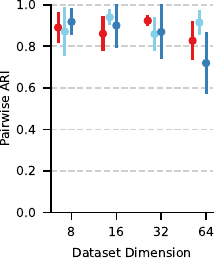}
        \caption{Seed Stability \\t-NEB}
        \label{fig:stability:tmm}
    \end{subfigure}%
    \begin{subfigure}[b]{0.23\linewidth}
        \includegraphics[height=3.6cm]{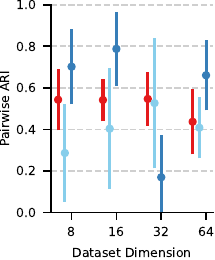}
        \caption{Seed stability \\ \qquad UniForCE}
        \label{fig:stability:uniforce}
    \end{subfigure}%
    \begin{subfigure}[b]{0.31\linewidth}
        \includegraphics[height=3.6cm]{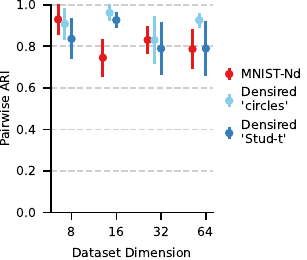}
        \caption{Overclustering stability \\t-NEB}
        \label{fig:stability:overclusteringTMM}
    \end{subfigure}
    \caption{Stability measured in pairwise ARI of the predictions (either against different seeds or against different number of initial components $m\in[20..55]$).}
    \label{fig:stability_uniforce}
\end{figure}

\paragraph{Stability against different seeds.}
Our method builds on mixture models which are sensitive to initialization and prone to getting stuck in local optima. 
To assess the stability of our method, we ran multiple trials with identical hyperparameters but varying random seeds and computed pairwise ARI values across ten initializations. 
Initialization had minimal impact, with pairwise ARI values mostly above 0.8 (\cref{fig:stability:tmm}), except for the Densired `Stud t' 64D, where stability was still good. 
We speculate that higher-dimensional spaces require more points for accurate density estimation, while our datasets have a fixed number of data points across dimensions (see \cref{tab:datasets}).

We repeated the experiment for SMMP \citep{guan2022smmp} and UniForCE \citep{vardakas2023UniForCE}. While SMMP is deterministic, UniForCE is clearly less stable, especially on the Densired `circles' dataset (\cref{fig:stability:uniforce}).

\begin{figure*}[t]
    \centering
    \includegraphics[width=\textwidth]{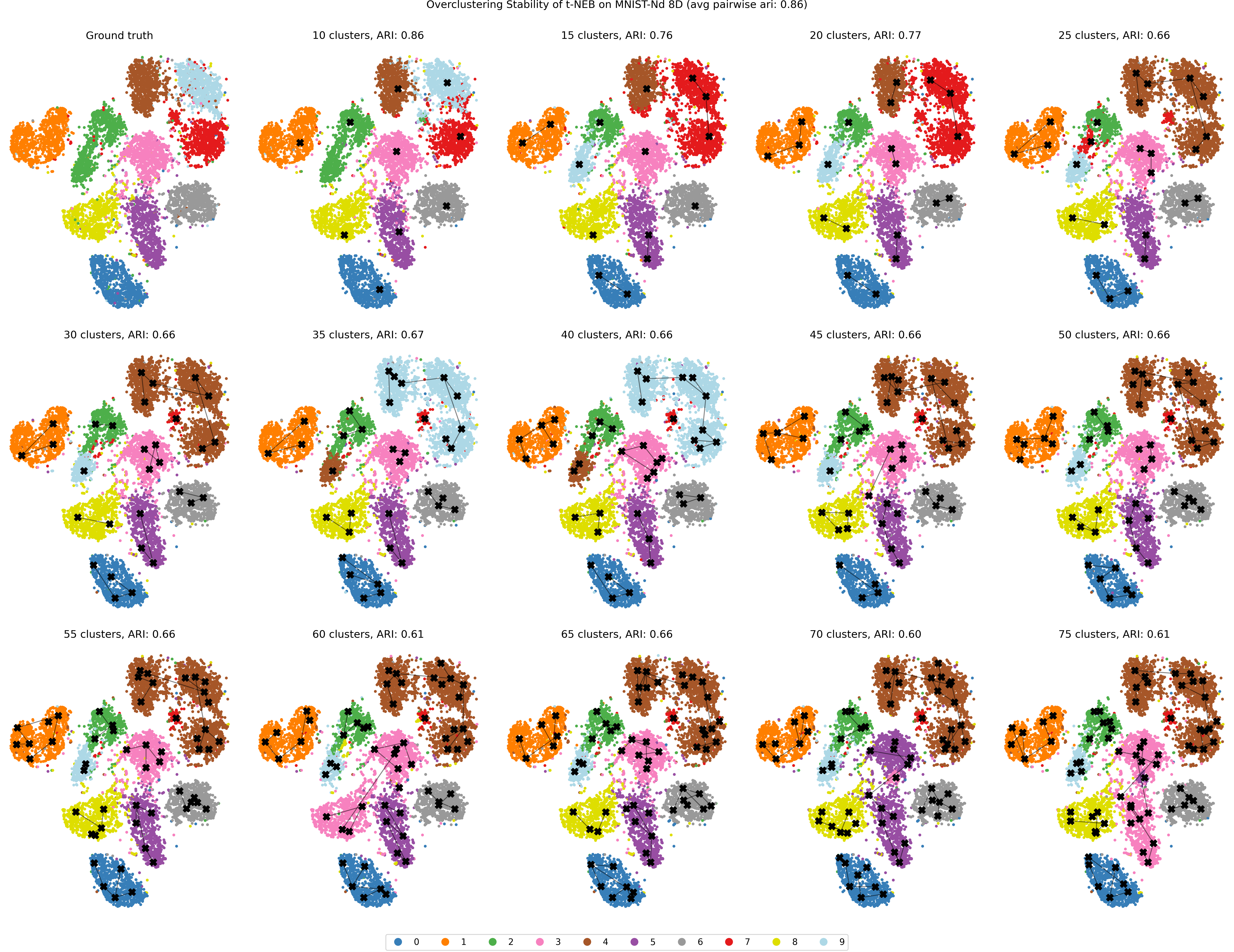}
    \caption{Overclustering stability of MNIST-Nd 8D. The model consistently splits the 2 while merging 7,4,9 which leads to low ARI. Clustering mistakes are mostly consistent across number of clusters in the density model.}
    \label{fig:overclustering_Mnist16}
\end{figure*}

\paragraph{Stability against number of initial components (Overclustering).}
\label{subsec:stability}
The main hyperparameter of our method is the degree of overclustering, i.e., the number of components in the initial mixture model. 
While different numbers of components result in different clusterings, it is unclear how much this impacts the final clustering produced by merging these components using NEB-based distances. 
We are thus interested in how similar the outcomes are after merging clusters.

In our experiment, we generate clusterings based on \( 10 + 5i \) mixture components with \( i \in [2..9] \). %
This range was chosen such that at least 10 merges are required to reach the target number of clusters and the maximum number of clusters is 55.
Pairwise ARI values between clusterings after merging to \( n \) clusters are shown in \cref{fig:stability:overclusteringTMM}.
Generally, higher clustering performance typically corresponds to higher stability, as less room for fluctuations remains.

Pairwise ARI of different initial components (\cref{fig:stability:overclusteringTMM}) are generally lower than those observed with random seeds (\cref{fig:stability:tmm}), which is expected since varying the overclustering level changes  both the amount of peaks and their position.
Overall, stability is only slightly worse than for the seed stability, indicating that clusters only marginally change when increasing the number of initial components during mixture model fitting.

We visualized clusterings with varying the number of initial components $m \in \{10, 15, \cdots 75\}$ on MNIST-Nd 8D using t-SNE (see \cref{fig:overclustering_Mnist16}).
The model consistently clusters the numbers 0, 1, 5 and 6 correctly.
Mistakingly, the 2 is split into two clusters several times, while 3 merges with 5 and 8, as well as 4, 7 and 9 are merged, leading to a relatively low ARI.
Overall, clustering results are mostly consistent across different initial number of clusters in the density model.

\section{Additional metrics for clustering evaluation}
\label{app:moremetrics}
While ARI is a common choice to measure the quality of clusterings in case the ground-truth data is available, we checked that this choice is indeed reasonable for our data and experiments.
We thus repeat our experiments and use Normalized Mutual Information (NMI) \citep{strehl2002cluster}, Fowlkes-Mallows (FM) \citep{fowlkes1983method}, and Variation of Information (VI) \citep{meilua2007comparing} as clustering quality assessment indexes for a more rigorous evaluation. 
Since all four metrics are highly correlated, we claim that our ARI-based conclusions are valid (\cref{tab:additional_metrics}).
\begin{table}
\caption{Clustering quality assessment metrics for various datasets and clustering methods. We can see that all metrics are highly correlated with each other. Best performing method in \textbf{bold}, second-best in \textit{italic}.}
\label{tab:additional_metrics}
\resizebox{\textwidth}{!}{
\begin{tabular}{lcccc|cccc|cccc}
\toprule
& \multicolumn{4}{c}{\textbf{Densired `circles'}} & \multicolumn{4}{c}{\textbf{Densired `Stud-t'}} & \multicolumn{4}{c}{\textbf{MNIST-Nd}} \\
 & 8D & 16D & 32D & 64D & 8D & 16D & 32D & 64D & 8D & 16D & 32D & 64D \\
\midrule \\
\multicolumn{13}{l}{\textbf{Adjusted Rand Index (ARI)} $\uparrow$} \\
\midrule
Agglomerative & 0.68 & 0.66 & 0.59 & 0.75 & 0.56 & 0.87 & 0.90 & 0.64 & 0.80 & 0.68 & 0.62 & 0.49 \\
HDBSCAN & 0.00 & 0.00 & 0.44 & 0.00 & 0.01 & 0.00 & 0.00 & 0.00 & 0.03 & 0.06 & 0.07 & 0.07 \\
Gaussian Mixture & 0.79 & \textit{0.77} & 0.69 & \textit{0.91} & 0.74 & 0.68 & 0.49 & 0.42 & \textbf{0.89} & 0.74 & 0.73 & 0.62 \\
Student-t Mixture & 0.81 & 0.69 & 0.58 & 0.61 & 0.74 & 0.90 & 0.90 & \textit{0.75} & 0.85 & 0.86 & \textit{0.90} & \textbf{0.77} \\
Leiden & \textit{0.83} & \textit{0.77} & \textit{0.76} & 0.89 & \textbf{0.89} & \textit{0.93} & \textit{0.91} & \textbf{0.80} & \textbf{0.89} & \textbf{0.92} & \textbf{0.93} & \textit{0.71} \\
t-NEB (ours) & \textbf{0.92} & \textbf{1.00} & \textbf{0.94} & \textbf{0.96} & \textbf{0.89} & \textbf{0.94} & \textbf{0.94} & 0.74 & 0.77 & \textbf{0.92} & 0.79 & 0.64 \\ \\

\multicolumn{13}{l}{\textbf{Normalized Mutual Information (NMI)} $\uparrow$} \\
\midrule
Agglomerative & 0.82 & 0.82 & 0.79 & 0.84 & 0.59 & 0.81 & 0.85 & 0.71 & 0.83 & 0.78 & 0.75 & 0.65 \\
HDBSCAN & 0.00 & 0.00 & 0.59 & 0.00 & 0.00 & 0.00 & 0.01 & 0.01 & 0.06 & 0.11 & 0.14 & 0.13 \\
Gaussian Mixture & 0.83 & \textit{0.90} & 0.84 & \textit{0.93} & 0.73 & 0.66 & 0.56 & 0.46 & \textbf{0.89} & 0.82 & 0.81 & 0.75 \\
Student-t Mixture & \textit{0.91} & 0.84 & 0.79 & 0.79 & 0.72 & 0.82 & 0.84 & 0.71 & 0.86 & 0.87 & \textit{0.89} & \textbf{0.84} \\
Leiden & \textit{0.91} & \textit{0.90} & \textit{0.90} & \textit{0.93} & \textbf{0.82} & \textbf{0.88} & \textit{0.86} & \textbf{0.81} & \textbf{0.89} & \textbf{0.91} & \textbf{0.91} & \textit{0.80} \\
t-NEB (ours) & \textbf{0.93} & \textbf{1.00} & \textbf{0.95} & \textbf{0.96} & \textbf{0.82} & \textit{0.87} & \textbf{0.88} & \textit{0.75} & 0.85 & \textbf{0.91} & 0.86 & 0.74 \\ \\

\multicolumn{13}{l}{\textbf{Fowlkes-Mallows (FM)} $\uparrow$} \\
\midrule
Agglomerative & 0.77 & 0.76 & 0.70 & 0.82 & 0.73 & 0.91 & 0.93 & 0.79 & 0.82 & 0.72 & 0.66 & 0.55 \\
HDBSCAN & 0.54 & 0.54 & 0.70 & 0.54 & 0.44 & 0.43 & 0.42 & 0.39 & 0.26 & 0.29 & 0.30 & 0.30 \\
Gaussian Mixture & 0.85 & \textit{0.84} & 0.78 & \textit{0.94} & 0.81 & 0.77 & 0.65 & 0.62 & \textbf{0.90} & 0.77 & 0.76 & 0.66 \\
Student-t Mixture & 0.87 & 0.78 & 0.69 & 0.71 & 0.81 & 0.93 & 0.93 & \textit{0.83} & 0.86 & 0.88 & \textit{0.91} & \textbf{0.79} \\
Leiden & \textit{0.88} & \textit{0.84} & \textit{0.83} & 0.92 & \textbf{0.92} & \textbf{0.95} & \textit{0.94} & \textbf{0.87} & \textbf{0.90} & \textbf{0.93} & \textbf{0.93} & \textit{0.76} \\
t-NEB (ours) & \textbf{0.94} & \textbf{1.00} & \textbf{0.96} & \textbf{0.97} & \textbf{0.92} & \textbf{0.95} & \textbf{0.95} & 0.82 & 0.80 & \textbf{0.93} & 0.82 & 0.67 \\ \\

\multicolumn{13}{l}{\textbf{Variation of Information (VI)} $\downarrow$} \\
\midrule
Agglomerative & 0.55 & 0.56 & 0.66 & 0.49 & 0.95 & 0.50 & 0.41 & \textit{0.65} & 0.78 & 1.00 & 1.12 & 1.57 \\
HDBSCAN & 1.41 & 1.42 & 0.82 & 1.41 & 1.94 & 1.97 & 1.97 & 2.07 & 2.80 & 2.65 & 2.59 & 2.60 \\
Gaussian Mixture & 0.51 & \textit{0.31} & 0.52 & 0.20 & 0.79 & 0.94 & 1.16 & 1.32 & \textbf{0.52} & 0.80 & 0.86 & 1.15 \\
Student-t Mixture & \textit{0.27} & 0.51 & 0.67 & 0.68 & 0.82 & 0.48 & 0.45 & 0.74 & 0.64 & 0.58 & \textit{0.50} & \textbf{0.73} \\
Leiden & \textit{0.27} & \textit{0.31} & \textit{0.32} & \textit{0.19} & \textbf{0.49} & \textbf{0.34} & \textit{0.37} & \textbf{0.47} & \textbf{0.52} & \textbf{0.41} & \textbf{0.40} & \textit{0.88} \\
t-NEB (ours) & \textbf{0.21} & \textbf{0.01} & \textbf{0.15} & \textbf{0.11} & \textit{0.50} & \textit{0.36} & \textbf{0.32} & 0.74 & 0.67 & \textit{0.43} & 0.63 & 1.16 \\
\bottomrule
\end{tabular}}
\end{table}

\section{Runtime}
In many cases, the fitting of the density model, especially in high-dimensional and large datasets is the bottleneck of this method, especially when using a GPU for the NEB computations.

Our NEB implementation is based on JAX and thus is able to exploit the parallelism of modern GPUs.
For TMM fitting, we used a CPU-based package, while implementing the EM-based fitting in JAX to make use of just-in-time compilation and accelerators. This is expected to speed up the fitting process massively, effectively removing the bottleneck.
We leave this engineering challenge for future work.

Experiments were run on a 10GB slice of an NVIDIA A100 plus 8 cores of a Zen3 EPYC 7513 and 64GB of memory.
Running entirely on CPU is also feasible, even though this will increase the share of NEB computations.

\section{Used Libraries}
\label{app:software}
\begin{itemize}[noitemsep]
    \item \citet{chan2019gpu} for t-SNE algorithm, implements t-SNE on CUDA
    \item  \citet{scikit-learn} for Scikit-learn for standard clustering algorithms
    \item \citet{wolf2018scanpy} SCANPY for PAGA implememtation
    \item \citet{DENSIRED} DENSIRED for creating high-dimensional non-Gaussian synthetic datasets
    \item  \url{https://github.com/jlparkI/mix_T} (studenttmixture) for fitting student-t mixture models
    \item \citet{jax2018github} for NEB computations
\end{itemize}

 \end{appendix}

\end{document}